\documentclass{article}
\pdfoutput=1

\usepackage{arxiv}

\usepackage[utf8]{inputenc} 
\usepackage[T1]{fontenc}    
\usepackage{hyperref}       
\usepackage{url}            
\usepackage{booktabs}       
\usepackage{amsfonts}       
\usepackage{nicefrac}       
\usepackage{microtype}      
\usepackage{lipsum}		
\usepackage{graphicx}

\usepackage[numbers]{natbib}
\usepackage{doi}
\usepackage{amsmath} 
\usepackage[utf8]{inputenc} 
\usepackage[english]{babel} 
\usepackage{amsmath}
\usepackage{amssymb}
\usepackage{txfonts}
\usepackage{mathdots}
\usepackage{graphicx}
\usepackage{algorithm}
\usepackage{algpseudocode}
\usepackage{algorithmicx}
\usepackage{varwidth}
\usepackage{url}
\usepackage{mathptmx}
\usepackage{caption} 
\usepackage{geometry}
\usepackage{graphicx}
\usepackage{subcaption}
\usepackage{float} 

\usepackage{bm}
\usepackage[nodisplayskipstretch]{setspace}
\usepackage{setspace} 

\usepackage{ragged2e}
\usepackage{titlesec}
\titlespacing{\section}{0.5pt}{\parskip}{-\parskip}
\titlespacing{\subsection}{0.5pt}{\parskip}{-\parskip}
\titlespacing{\subsubsection}{0.5pt}{\parskip}{-\parskip}

\title{Reservoir History Matching of the \textit{Norne} Field with generative exotic priors and a coupled Mixture of Experts - Physics Informed Neural Operator Forward Model}

\author{
  Clement Etienam \\
  Nvidia \\
  \texttt{cetienam@nvidia.com} \\
   \And
  Yang Juntao \\
  Nvidia \\
  \texttt{yjuntao@nvidia.com} \\
   \And
  Oleg Ovcharenko \\
  Nvidia \\
  \texttt{oovcharenko@nvidia.com} \\
   \And
  Issam Said \\
  Nvidia \\
  \texttt{Isaid@nvidia.com} \\
}

\date{}

\renewcommand{\headeright}{}
\renewcommand{\undertitle}{}

\newcommand{\by}{{\mathbf y}}
\newcommand{\bx}{{\mathbf x}}
\newcommand{\bz}{{\mathbf z}}
\newcommand{\bof}{{\mathbf f}}

\newcommand{\cO}{\mathcal{O}}

\begin{document}
\maketitle
\begin{abstract}
We have developed a novel reservoir characterization workflow that solves the reservoir history matching problem coupling a physics Informed neural operator (\textit{PINO}) forward model with a mixture of experts’ approach termed cluster classify regress (\textit{CCR}). The inverse modelling is computed via an adaptive Regularized Ensemble Kalman inversion (\textit{a}REKI) method, and the method is optimal for rapid inverse uncertainty quantification tasks during reservoir history matching. We parametrize the unknown permeability and porosity fields for the case of sampling non-gaussian posterior measures by using, for instance, a variational convolution autoencoder and another instance using a denoising diffusion implicit model (DDIM). The \textit{CCR} serves as exotic priors in the inverse modelling and as a coupled supervised model with the prior \textit{PINO} surrogate for replicating the nonlinear Peaceman well equations in the forward model. The \textit{CCR} is flexible, and any separate and independent machine learning algorithm can be used for each of the 3 stages. The loss function required for the \textit{PINO} reservoir surrogate updating is obtained from the combined supervised data loss and losses arising from the initial conditions and residual of the governing black oil pde. The output from the coupled  \textit{PINO-CCR} surrogate is the pressure, water and gas saturations as well as the well outputs of oil, water, and gas production rates. The developed methodology is first compared to a standard numerical black oil reservoir simulator for a waterflooding case on the \textit{Norne} field, and the similarity of the outputs with those from the \textit{PINO-CCR} surrogate is very close. Next, we use this developed \textit{PINO-CCR} surrogate in the developed \textit{a}REKI history matching workflow. The overall workflow is successful in recovering the unknown permeability and porosity field and simulation is very fast with a speedup of up to 6000X to conventional numerical reservoir simulation approaches. Learning the \textit{PINO-CCR} surrogate on an NVIDIA H100 with 80G memory takes approximately  5 hours for 100 training samples of the \textit{Norne} field. The workflow is also suitable to be used in an ensemble-based workflow, where sampling the posterior density, given an expensive likelihood evaluation, to quantify uncertainty, as with most scientific computation, is desirable.
\end{abstract}
\section{Introduction.}
\subsection{Overview}
\noindent Reservoir engineers are mostly burdened with the task of developing reliable simulation models that can accurately capture the complex dynamics of subsurface reservoir systems. The use of a set of observed measurements, $\boldsymbol{d}$ , to infer a set of model parameters, $\boldsymbol{m}$ , is known as an inverse problem  \cite{emerick2013ensemble,stuart2010inverse,tarantola2005inverse,oliver2010recent,nocedal1999numerical,iglesias2013regularizing}.~ Given specific observations which are the observed data of the state variable (data that is a direct function of state variables); the aim is to infer information about the model parameters which includes some uncertainty. The focus is on discrete inverse problems (systems characterized by a few or a finite number of parameters). Reservoir history matching is an example of a discrete inverse problem. In such situations, the number of data is always finite, and the data always contain some measurement errors \cite{emerick2013ensemble,etienam2024novel,oliver2010recent,aanonsen2009ensemble,chen2011ensemble}. It is hence impossible to estimate the parameters of a reservoir model correctly from inconsistent, inaccurate and insufficient data using a forward model that most likely will contain some modeling errors. In most cases, we always have some prior information about the realism and plausibility of the reservoir model, these include a geological model that is constructed from cores, logs and seismic data and also some information on the depositional environment.

\noindent Reservoir history matching is the process of constraining the reservoir model to mimic actual historical dynamic production data of the reservoir to predict its future production behavior and reduce uncertainty about the description of the reservoirs \cite{emerick2013ensemble,etienam2024novel,oliver2010recent,aanonsen2009ensemble,chen2011ensemble}. Reservoir history matching is an ill-posed problem because the amount of independent data available is much less than the number of variables. Hence, there exist an infinite number of combinations of the unknown reservoir properties that result in reservoir models which can match the observations. Besides that, the information accessible about the reservoir is always inaccurate and may be inconsistent \cite{emerick2013ensemble,etienam2024novel,oliver2010recent,aanonsen2009ensemble,chen2011ensemble}. As a result, reservoir models are constructed with uncertain parameters; consequently, their predictions are also uncertain.

\noindent ~Over the last decade, increased importance has been attributed to the quantification of uncertainty in reservoir model performance predictions and reservoir model descriptions to manage risk. Because of this obsession with the characterization of uncertainty, it is now common to generate multiple history-matched models. However, generating multiple history-matched models may not necessarily lead to a correct assessment of uncertainty. Uncertainty has no scientific meaning beyond the realm of statistics and probability. Bayesian statistics provides a straightforward theory for dealing with uncertainty.

\noindent ~In the application of interest, Bayes' theorem allows one to write down the posterior probability density function (pdf) for reservoir model parameters conditional to field measurements such as production, electromagnetic and seismic data. Then, the optimization problem of characterizing the model uncertainty in reservoir model parameters is tantamount to the problem of sampling this posterior pdf \cite{stuart2010inverse,tarantola2005inverse,law2012evaluating}. If a set of realizations of the vector of reservoir model parameters represents a set of samples from the posterior pdf, then a correct assessment of the uncertainty in specific outcomes of reservoir performance predictions can be generated by predicting with each model and then constructing statistics for the set of outcomes. For example, one can estimate the pdf for a well's predicted oil rate at each time from a histogram of the predictions set. Markov chain Monte Carlo (MCMC) provides a theoretically attractive technique for sampling the posterior pdf for model parameters.

\noindent ~It is established that a properly designed MCMC method samples this pdf correctly in the limit as the number of states in the chain goes to infinity. For high-dimensional problems, MCMC typically requires many iterations to provide a reasonable sampling of the desired pdf. When utilizing MCMC to compute the posterior pdf of reservoir model parameters conditional to historical observed production data, computation of the probability of accepting the transition from the current state to the proposed state requires a run of the reservoir model simulator to evaluate the likelihood part associated with the posterior pdf. This requirement makes the direct application of MCMC to realistic reservoir problems computationally prohibitive \cite{stuart2010inverse,tarantola2005inverse,law2012evaluating}.

\noindent ~The modeling of flow and transport processes in subsurface formations is essential for various reservoir engineering applications, such as oil reservoir development and carbon sequestration. Quantifying flow mechanisms in these scenarios often requires using partial differential equations (PDEs) founded on laws of conservation. Traditionally, PDEs have been solved numerically using methods such as finite elements, finite differences, spectral, and mesh-less techniques. These methods have undergone significant development over the years, rendering them robust and flexible for solving complex subsurface flow and transport problems. However numerical simulations can be computationally demanding for large-scale simulation problems in petroleum engineering, posing challenges during execution \cite{wang2022,OPM,rojas1998nonlinear}. 

\noindent ~ Moreover, tasks like history-matching, sensitive analysis, and project design optimization require multiple simulation runs, leading to prohibitively inefficient computations for obtaining meaningful results  \cite{Ertekin2019}.In recent years, data-driven methods have garnered significant attention and achieved remarkable progress in various domains such as natural language processing \cite{otter2021} and image classification\cite{Lecun2015}. Among these methods, artificial neural networks (ANNs) have emerged as one of the most important data-driven simulation techniques due to their universal approximation property and ability to accurately approximate any measurable function. The versatility of ANNs extends to various science and engineering fields, demonstrating their potential in nonlinear universal approximation and data assimilation \cite{etienam2024novel}.Notably, within the realm of petroleum engineering, ANNs have made substantial advancements in both forward modeling and other applications. Dong et al.\cite{Dong2019} proposed an enhanced artificial neural network model for predicting CO2 minimum miscibility pressure based on the full composition of the crude oil and temperature.

\noindent ~ Erofeev et al.\cite{Erofeev2019} studied the applicability of various machine learning algorithms for predicting rock properties usually defined by geoscientists through specialized laboratory analysis. Moosavi et al. \cite{Moosavi2019} utilized multilayer perceptron neural networks to accurately identify reservoir models from pressure derivative curves derived from horizontal wells. Chung et al. \cite{chung2020} proposed a pore-scale finite volume solver to predict permeability on digital cores by solving flow on micro-CT images. Kim et al. \cite{kim2021} proposed an innovative data-integration method that uses an iterative-learning approach with a deep neural network coupled with a stacked autoencoder to address challenges encountered in many-objective history matching. Santos et al. \cite{santos2021} tackled the limitation of modeling important geometries like fractures and vuggy domains accurately using a general multiscale deep learning model that can learn from porous media simulation data. Furthermore, Wang et al. \cite{wang2022} outlined an integrated method that combines predictions of fluid flow with direct flow simulation, significantly reducing computation time without compromising accuracy. Alakeely and Horne \cite{Alakeely2022} examined the effectiveness of generative deep learning methods in predicting multiphase flow profiles of new wells in unseen locations using historical production data and a variational autoencoder algorithm. 

\noindent ~Dong et al. \cite{Dong2019} introduced a deep reinforcement learning based approach for automatic curve matching for well test interpretation, utilizing the double deep Q-network. Despite its remarkable advancements, the conventional data-driven approach inevitably faces several challenges. Firstly, the data-driven model is perceived as a "black box," since it lacks the incorporation of physical meaning of the dataset, leading to predictions that may be physically inconsistent or implausible\cite{raissi2019physics}. Secondly, the robustness of the data-driven model maybe poor, and its long-term prediction capabilities are weak. This can be attributed to the fact that generalizing and extrapolating beyond the parameter space of the included dataset is remain extremely challenging, as the models developed within artificial neural networks are bounded by the parameter space of the training dataset \cite{raissi2019physics,Kania2021}. A recent and innovative neural network architecture that embodies this concepts known as 'physics-informed neural networks’ (PINNs)\cite{raissi2019physics,Kania2021}. By incorporating PDEs, boundary conditions, initial conditions, and other measurable prior knowledge into the loss function, PINNs seamlessly integrate data and physics laws to construct a physics-constrained loss function based on automatic differentiation. These algorithms not only improve the interpretability of machine learning models in terms of physical principles but also demonstrate robustness in handling flawed data, such as missing or noisy values, outliers, and other irregularities \cite{raissi2019physics,Kania2021}. 

\noindent ~The performance of PINNs has been demonstrated in petroleum reservoir problems associated with subsurface flow and transport in porous media. Li et al. \cite{Li2022} introduced a TGNN model that integrates high-fidelity numerical simulation data, fundamental physical laws, boundary conditions, initial conditions, expert knowledge, and engineering control terms in the loss function, enabling the solution of two-phase subsurface f low problems at the reservoir scale beyond the Buckleye Leverett equation. Daolun et al. \cite{Daolun2021} proposed an improved physics constrained PDE solution method by integrating potential features of the PDE into the loss functions to alleviate the strong nonlinear problem of the flow equation caused by the source-sink term in the single-phase homogeneous reservoir problem. Gasm and Tchelepi (2021) \cite{Gasmi2021} utilized PINNs to tackle the two-phase immiscible flow problem governed by the BuckleyeLeverett equation, achieving physical solutions by incorporating either a diffusion term into the partial differential equations or observed data. Almajid and Abu-Al-Saud \cite{Alma2022} implemented a physics informed neural network technique that combines fluid flow physics and observed data to model the Buckleye Leverett problem. 

\noindent ~Cornelio et al. \cite{cornelio2022} developed a neural network (NN) model to learn the physical model residual errors in simulation-based production prediction as a function of input parameters of an unconventional well. Hanna et al. \cite{hanna2022} developed a novel residual based adaptive PINN and compared it with the residual-based adaptive refinement (RAR) method and a PINN with fixed collocation points. Wang et al. \cite{wang2022} proposed a theory-guided convolutional neural network (TgCNN) framework that incorporates discretized governing equation residuals into the training of convolutional neural networks, to extend to two-phase porous media f low problems. However, when applying PINNs to reservoir simulation with large-scale systems and production wells, existing research typically relies on training neural networks using high-fidelity numerical simulation data. The challenge arises from the scarcity of real reservoir data available for training purposes, which is often limited to information such as production rates and bottomhole flow pressures. Consequently, the accuracy of the physical information method may not always be guaranteed, when solving complex partial differential equations (PDEs) with large-scale systems and production wells, especially in the near-well zone where the pressure gradient is the largest. In this study, we extend the methodology proposed by Etienam et al. \cite{etienam2024novel},  focusing entirely on the \textit{Norne} field \cite{Norne1} by incorporating denoising diffusion implicit model\cite{song2020denoising} and variational convolutional autoencoder  \cite{kingma2014autoencoding} as generative exotic priors, we construct a forward solver surrogate that couples a modified mixture of experts approach \cite{etienam2023ultrafast, bernholdt2019cluster} with the \textit{PINO} to give an augmented surrogate we call \textit{PINO-CCR}. We then use the \textit{PINO-CCR} surrogate in a Bayesian ensemble history matching workflow to recover the unknown permeability,porosity and fault multipliers parameters. 
Our contribution includes,
\begin{enumerate}
\item \textbf{ }Develop a Fourier neural operator surrogate (\textit{PINO}-surrogate) trained from both the data and physics loss to replace the black oil model reservoir equations with Nvidia Modulus framework.\textbf{}
\item \textbf{ }Develop an inverse modeling workflow using the prior developed PINO surrogate together with exotic priors to calibrate the permeability and porosity field for the reservoir under consideration.\textbf{}
\item \textbf{ }Use the developed inverse modeling workflow for forward uncertainty quantification and field optimization.
\end{enumerate}
\section{Forward problem}
\noindent Our simplified Black oil model for three-phase flow in porous media for reservoir simulation is given as \cite{dorn2008history2,rojas1998nonlinear}.
\begin{equation}
\varphi \frac{\partial S_w}{\partial t} - \nabla \cdot \left[T_w\left(\nabla P_w + \rho_w g k\right)\right] = Q_w\tag{1a}
\label{eq:1a}
\end{equation}
\begin{equation}
\varphi \frac{\partial S_o}{\partial t} - \nabla \cdot \left[T_o\left(\nabla P_o + \rho_o g k\right)\right] = Q_o\tag{1b}
\label{eq:1b}
\end{equation}
\begin{equation}
\nabla \cdot \left(\frac{\rho_g}{B_g} T_g \left(\nabla P_g + \rho_g g k\right) + \frac{R_{so} \rho_g}{B_o} T_o \left(\nabla P_o + \rho_o g k\right)\right) - Q_g = -\frac{\partial}{\partial t}\left[\varphi \left(\frac{\rho_g}{B_g} S_g + \frac{R_{so} \rho_g}{B_o} S_o\right)\right]\tag{1c}
\label{eq:1c}
\end{equation}
\noindent subscript \(w = \text{water}, o = \text{oil}, g = \text{gas}\). The system is closed by adding three additional equations \(P_{cwo} = P_o - P_w; P_{cog} = P_g - P_o; S_w + S_o + S_g = 1\). Gravity effects are considered by the terms \(\rho_w g k\) and \(\rho_o g k\). \(\Omega \subset \mathbb{R}^n\) (\(n = 2, 3\)) is the modeling domain with boundary \(\partial \Omega\), and \([0, t_f]\) is the time interval for which production data is available. \(\varphi(x)\) = porosity, \(T_o, T_w, T_g\) and \(T\) the transmissibilities,
\begin{equation}
T_w = \frac{K(x) K_{rw}(S_w)}{\mu_w}; T_o = \frac{K(x) K_{ro}(S_w)}{\mu_o}; T_g = \frac{K(x) K_{rg}(S_o)}{\mu_g};\tag{2}
\label{eq:2}
\end{equation}
Assuming fault transmissibility multiplier (FTM),
\begin{equation}
T = FTM \cdot (T_w + T_o + T_g)\tag{3a}
\label{eq:3a}
\end{equation}
\begin{equation}
- \nabla \cdot \left[T \nabla P\right] = Q\tag{3b}
\label{eq:3b}
\end{equation}
\begin{equation}
\varphi \frac{\partial S_w}{\partial t} - \nabla \cdot \left[T_w \left(\nabla P\right)\right] = Q_w\tag{3c}
\label{eq:3c}
\end{equation}
\begin{equation}
\nabla \cdot \left(\frac{\rho_g}{B_g} T_g \left(\nabla P\right) + \frac{R_{so} \rho_g}{B_o} T_o \left(\nabla P\right)\right) - Q_g = -\frac{\partial}{\partial t}\left[\varphi \left(\frac{\rho_g}{B_g} S_g + \frac{R_{so} \rho_g}{B_o} S_o\right)\right]\tag{3d}
\label{eq:3d}
\end{equation}
\noindent Boundary and initial conditions are given as \(S_w(x,0) = S^0_w(x); P(x,0) = P^0_w(x); \nabla p \cdot \nu = 0\). \(\nu\) = outward unit normal to \(\partial \Omega\).
\noindent The overall elliptical-hyperbolic coupled PDE equations to solve are equations \(\ref{eq:3b}\), \(\ref{eq:3c}\), and \(\ref{eq:3d}\).
\subsection{Surrogate forward modeling with a Fourier Neural operator (\textit{FNO}) - Contribution}
\noindent If the notation above is clear, we now discuss the \textit{PINO} model used for solving the forward problem posed in Eqn.3(b-d).
\begin{figure}[h!]
\centering
\includegraphics[width=5.74in, height=1.35in]{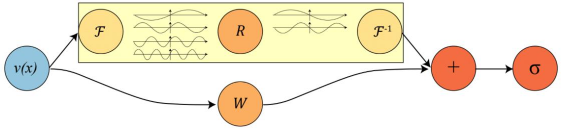}
\caption{\ \textit{A standard Fourier neural operator (FNO) as adopted from \cite{li2021fourier}}}
\end{figure}
The Fourier neural operator (\textit{FNO}) is effective in learning the mapping relationships between the infinite-dimensional function spaces that convert a traditional convolution operation into a multiplication operation using the Fourier transform \cite{li2021fourier}. This causes the higher modes of the neural layer to be removed from the Fourier space, leaving only the lower modes in the Fourier layer. Hence, \textit{FNO} can greatly improve the computational efficiency of the training process. The goal of \textit{FNO} is to use a neural network $G^{\dagger }(a)\ =\ u$ to approximate the mapping of function $G\ :\ A\ \to \ U,\ G\ \approx \ G^{\dagger }$.
\begin{equation}
e_{n+1}(x)\ =\ \sigma (We_n(x)\ +\ (\wp (a;\ \vartheta )e_n(x))(x)) \tag{4a}
\label{eq:4a}
\end{equation}
Figure 1 shows the full \textit{FNO} architecture; input observation $v(x)$ is first lifted to a higher dimensional potential representation $e_0(x)\ =\ M(v(x))$ by a fully connected shallow neural network $M$. Then, $e_0(x)\ $is used as input for an iterative Fourier layer $\ e_0\to \ e_1\ \to \ e_2\ \to \ \textrm{·}\ \textrm{·}\ \textrm{·}\ \to \ e_N$. The concrete iterative process from state $e_n\ \to \ e_{n+1}$ can be written as:
\begin{equation}
e_{n+1}(x)\ =\ \sigma (We_n(x)\ +\ (\wp (a;\ \vartheta )e_n(x))(x)) \tag{4b}
\end{equation}
Figure 1. Architecture of the \textit{FNO}. $\digamma $ denotes the Fourier transform, and ${\digamma }^{-1}$ denotes the inverse Fourier transform. $R_{\vartheta }$ is a Fourier transform of the periodic function $\wp $; $W$ is a local linear transform; $\sigma $ is a non-linear activation function; $e_0$ and $e_1$ are the input and output of Fourier layer 1. Here, $\sigma $ is a non-linear activation function, $W$ is a linear transformation, and $\wp $ is an integral kernel transformation parameterized by a neural network. $(\wp (a;\ \vartheta )e_n(x))(x)$ is the bias, which can be written as an integral:
\begin{equation}
\left(\wp \left(a;\ \vartheta \right)e_n\left(x\right)\right)\left(x\right)=\ \int{{\wp }_{\vartheta }}\left(x,y\right)e_n\left(y\right)dy \tag{4c}
\end{equation}
To accelerate the integration process $\int{{\wp }_{\vartheta }}\left(x,y\right)e_n\left(y\right)dy$, Li et al. (2020) imposed a condition that ${\wp }_{\vartheta }\left(x,y\right)=\ {\wp }_{\vartheta }(x-y)$ and the new integral becomes,
\begin{equation}
\left(\wp \left(a;\ \vartheta \right)e_n\left(x\right)\right)\left(x\right)=\ \int{{\wp }_{\vartheta }}\left(x,y\right)e_n\left(y\right)dy={\digamma }^{-1}(R_{\vartheta }.\left(\digamma e_n\right))(x) \tag{4d}
\end{equation}
\noindent where $\digamma $ denotes the Fourier transform, and ${\digamma }^{-1}$denotes the inverse Fourier transform. $R_{\vartheta }$ is a Fourier transform of periodic function $\wp $, $R_{\vartheta }\ =\ \digamma (\wp \vartheta )$, and $\wp \vartheta $ is the kernel of the neural network. Following the $N$ Fourier layer iteration, the last state output $e_N(x)$ is projected back to the original space using a fully connected neural network $S,\ y(x)\ =\ S(e_N(x)\mathrm{\ }).$
\noindent In every block of the Fourier layer, \textit{FNO} approximates highly non-linear functions mainly via combining the linear transform $W$ and global integral operator $R_{\vartheta }$ (Fourier transform). $R_{\vartheta }$can be parameterized as a complex-valued tensor in the Fourier space from truncating high Fourier frequency modes $k$, and it can greatly reduce the number of trainable parameters and improve efficiency during the process of training.
\noindent 
\noindent We are interested in predicting the pressure, water \& gas saturation given any input of the absolute permeability field ($K$) and effective porosity ($\varphi $). For completeness, the input and output tensors are indicated below.
$P_{ini},S_{ini}$ are the initial pressure and water saturation, $dt$ is the change in time steps.
Three different \textit{FNO} models are trained simultaneously for the pressure, gas saturation and water saturation, with the same interacting loss function, this is to accommodate for the difference in magnitude between these quantities.
\begin{equation}
X = \left\{K, \varphi , FTM, P_{ini},S_{ini}\right\}\in {\mathbb{R}}^{B_0\times 1,D\times W\times H}\tag{4e}
\end{equation}
\begin{equation}
P=f_1\left(X;{\theta }_p\right),\ S_w=f_2\left(X;{\theta }_s\right),S_g=f_3(X;{\theta }_g)\tag{4f}
\end{equation}
$K${ is the absolute permeability,  $Q$ is the total flow, $Q_{w}$  is the water flow (injection rate), $Q_{g}$  is the gas flow (injection rate), $dt$ is the time stepping, $P_{ini}$ is the initial pressure of the reservoir, and $S_{ini}$ is the initial water saturation of the reservoir.$P$ is the pressure, and $S_{w}$ is the water saturation and $S_{g}$ is the gas saturation 
$f_{1}(X;{\theta }_p)$ is the \textit{FNO} model for the pressure output,$f_{2}(X;{\theta }_s) $ is the \textit{FNO} model for the water saturation output, and $f_{3}(X;{\theta }_s)$ is the \textit{FNO} model for the gas saturation output
The physics loss ansatz is then,
\begin{equation}
{V\left(Q,p\mathrm{;}T\right)}_{pressure~equation}=\frac{1}{n_s}\left(~{\left\|\ \left(-~\nabla ~.~\left[T\nabla P\right]-Q\right).a_{ct}\right\|}^{\mathrm{2}}_{\mathrm{2}}\right)\mathrm{\ }\tag{5a}
\end{equation}
\begin{equation}
{V\left(p,S_w\mathrm{;}t,T_w\right)}_{\mathrm{water}~equation}=\frac{1}{n_s}{\left\|\left(\left(\varphi \frac{\partial S_w}{\partial t}-~\nabla ~.~\left[T_w.FTM.\left(\nabla P\right)\right]\right)-Q_w\right).a_{ct}\mathrm{\ }\right\|}^{\mathrm{2}}_{\mathrm{2}}\tag{5b}
\end{equation}
\begin{equation}\tag{5c}
\begin{split}
&V(p, S_g; t, T_g)_{\text{gas equation}} = \\
&\frac{1}{n_s}\left\| \nabla \cdot \left(\frac{\rho_g}{B_g}T_g \cdot \text{FTM}(\nabla P) + \frac{R_{so}\rho_g}{B_o}T_o \cdot \text{FTM}(\nabla P)\right) - Q_g + \frac{\partial}{\partial t} \left[\varphi \left(\frac{\rho_g}{B_g}S_g + \frac{R_{so}\rho_g}{B_o}S_o\right) \cdot a_{ct}\right] \right\|^{2}_{2}
\end{split}
\end{equation}
\begin{equation}
{\phi }_{cfd}={V\left(Q,p\mathrm{;}T\right)}_{pressure~equation}~+~{V\left(p,S_w\mathrm{;}t,T_w\right)}_{\mathrm{water}~equation}~+~{V\left(p,S_g\mathrm{;}t,T_g\right)}_{\mathrm{gas}~equation}\tag{5d}
\end{equation}
Together with the initial conditions and boundary conditions.
\begin{equation}
S_w\left(x,0\right)\mathrm{=}~S^0_w\left(x\right)~~\mathrm{;}~P\left(x,0\right)\mathrm{=}~P\left(x\right)~\mathrm{;}S_g\left(x,0\right)\mathrm{=}~S^0_g\left(x\right){;~F}_y\mathrm{=0}\tag{6a}
\end{equation}
\begin{equation}\tag{6b}
{\phi }_{ic}\mathrm{=}\frac{1}{n_s}{\left\|S_w\left(x,0\right)-~S^0_w\left(x\right)\right\|}^{\mathrm{2}}_{\mathrm{2}}\mathrm{+}\frac{1}{n_s}{\left\|P\left(x,0\right)-P\left(x\right)\right\|}^{\mathrm{2}}_{\mathrm{2}}\mathrm{+}\frac{1}{n_s}{\left\|S_g\left(x,0\right)-~S^0_g\left(x\right)\right\|}^{\mathrm{2}}_{\mathrm{2}}
\end{equation}
\begin{equation}\tag{6c}
{\phi }_{bc}\mathrm{=}~\frac{1}{n_s}{\left\|F_y\right\|}^{\mathrm{2}}_{\mathrm{2}}
\end{equation}
\begin{equation}\tag{6d}
\boldsymbol{\phi }\boldsymbol{\mathrm{=}}~{\boldsymbol{\phi }}_{\boldsymbol{cfd}}\boldsymbol{\mathrm{+}}{\boldsymbol{\phi }}_{\boldsymbol{ic}}\boldsymbol{\mathrm{+}}{\boldsymbol{\phi }}_{\boldsymbol{bc}}\boldsymbol{+}{\boldsymbol{\phi }}_{\boldsymbol{data}}
\end{equation}
\begin{equation}\tag{6e}
\theta \mathrm{=}~{\left[{\theta }_p,{\theta }_s,{\theta }_g\right]}^T
\end{equation}
\begin{equation}\tag{6f}
{\theta }^{j\mathrm{+1}}\mathrm{=}{\theta }^j-{\epsilon \nabla \phi }^j_{\theta }
\end{equation}
\begin{algorithm}[H]
\small
\setstretch{0.1} 
  \caption{PINO Reservoir Simulator Surrogate for Dynamic Properties}
\small
 \label{alg1}
  \begin{algorithmic}[1]
    \Require
    \begin{itemize}
      \item $X_1 = \{K, \varphi, FTM, P_{\text{ini}}, S_{\text{ini}}\} \in \mathbb{R}^{B_1 \times 1 \times D \times W \times H}$\Comment{dictionary of varied tenosr inputs}
      \item $X_{N1} = \{Q_w, Q_g, Q, dt, a_{ct}\} \in \mathbb{R}^{B_1 \times T \times D \times W \times H}$\Comment{dictionary of fixed tensor inputs}
      \item $Y_{1pt}$ -- labelled pressure,  $Y_{1st}$ -- labelled water saturation, $Y_{1gt}$ -- labelled gas saturation, $f_1(:, \theta_p)$, $f_2(:, \theta_s)$,$f_3(:, \theta_g)$
     \item $T$\Comment{Time reporting steps}
      \item $epoch$, $tol$, $w_1$, $w_2$, $w_3$, $w_4$, $w_5$, $w_6$ $\in \mathbb{R}$
    \end{itemize}      
    \Ensure
      \State $j = 0$
      \While{$(j \leq \text{epoch})$ or $(\phi \leq \text{tol})$}
        \State $Y_{1p} = f_{1}(X_1; \theta_p) \cdot a_{ct}$
        \State $Y_{1s} = f_{2}(X_1; \theta_s) \cdot a_{ct}$
        \State $Y_{1g} = f_{3}(X_1; \theta_g) \cdot a_{ct}$           
        \State \textbf{ Compute}: $T_w = \frac{K(x) \cdot K_{rw}(Y_{1s})}{\mu_w}$
        \State $T_o = \frac{K(x) \cdot K_{ro}(1 - (Y_{1s} + Y_{1g}))}{\mu_o}$
        \State $T_g = \frac{K(x) \cdot K_{rg} \cdot Y_{1g}}{\mu_g}$
        \State $T = X_{N1}[a_{ct}] \cdot X_{N1}[FTM] \cdot (T_w + T_o + T_g)$       
        \State \textbf{ Compute}: $V(Y_{1p}, Y_{1s}; t, T_w)_{1s} = \frac{1}{n_s} \cdot \left\|(\varphi \cdot \frac{\partial Y_{1s}}{\partial t} - \nabla \cdot [T_w \cdot \nabla Y_{1p}] - X_1[Q_w])\right\|^2_2$        
        \State \textbf{ Compute}:
	\begin{varwidth}[t]{\linewidth}
	\begin{align*}V(Y_{1p}, Y_{1s}, Y_{1g}; t, T_o, T_g)_{1g} =\frac{1}{n_s} \cdot & \left\| \nabla \cdot \left( \frac{\rho_g}{B_g} \cdot T_g \cdot X_{N1}[a_{ct}] \cdot X_{N1}[FTM] \cdot \nabla Y_{1p} \right. \right. \\
	& \left. \left. + \frac{R_{so} \cdot \rho_g}{B_o} \cdot T_o \cdot X_{N1}[a_{ct}] \cdot X_{N1}[FTM] \cdot \nabla Y_{0p} \right) - X_{N0}[Q_g] \right. \\
	& \left. + \frac{\partial}{\partial t} \left( \varphi \cdot \left( \frac{\rho_g}{B_g} \cdot Y_{1g} + \frac{R_{so} \cdot \rho_g}{B_o} \cdot (1 - (Y_{1s} + Y_{1g})) \right) \right) \right\|^2_2
	\end{align*}
	\end{varwidth}        
        \State \textbf{ Compute}: $V(X_{1}[Q], Y_{1p}; T)_{1p} = \frac{1}{n_s} \cdot \left\|(-\nabla \cdot [T \cdot a_{ct} \cdot \nabla Y_{1p}] - X_{1}[Q])\right\|^2_2$        
        \State $\phi_p = \left\|Y_{1pt} - f_{1}(X_{1}; \theta_p)\right\|^2_2$ \Comment{data supervised loss for pressure}
        \State $\phi_s = \left\|Y_{1st} - f_{2}(X_{1}; \theta_s)\right\|^2_2$ \Comment{data supervised loss for water saturation}
        \State $\phi_g = \left\|Y_{1gt} - f_{3}(X_{1}; \theta_g)\right\|^2_2$ \Comment{data supervised loss for gas saturation}       
        \State $\phi = w_1 \cdot \phi_p + w_2 \cdot \phi_s + w_3 \cdot \phi_g  + w_4 \cdot V(Y_{1p}, Y_{1s}; t, T_w)_{1s} +
 w_5 \cdot V(X_{1}[Q], Y_{1p}; T)_{1p} + w_6 \cdot V(Y_{1p}, Y_{1s}, Y_{1g}; t, T_o, T_g)_{1g}$\Comment{overall loss}        
        \State \textbf{Update models:}
        \State $\theta = [\theta_p, \theta_s, \theta_g]^T$
        \State $\theta^{j+1} = \theta^j - \epsilon \nabla \phi^j_{\theta}$\Comment{Back-propagation}        
        \State $j \leftarrow j + 1$
      \EndWhile      
    \State \textbf{Output:}:  $f_1(:, \theta_p)$, $f_2(:, \theta_s)$,  $f_3(:, \theta_g)$\Comment{pressure,water saturation and gas saturation \textit{PINO} surrogates}
  \end{algorithmic}
\end{algorithm}$w_{\mathrm{1}}{\dots ,w}_{\mathrm{6}}$ are the weights associated to the loss functions associated to the 6 terms. $X_{\mathrm{1}}\mathrm{=}\left\{K\mathrm{,\ FTM,}\varphi ,P_{ini},S_{ini}\right\}\mathrm{\ }$ are the dictionary inputs from running the reservoir simulator. $epoch,tol$ are the number or epochs and the tolerance level for the loss function. $f_{\mathrm{1}}\left(\mathrm{:,}{\theta }_p\right),f_{\mathrm{2}}\left(\mathrm{:,}{\theta }_s\right)\mathrm{,\ }f_3\left(\mathrm{:,}{\theta }_g\right)$ are the \textit{FNO} models for the pressure, water, and gas saturation equation respectively,\textit{ }$a_{ct}$ is the inactive /active cells grid, \textit{FTM} is the fault transmissibility multiplier.
\section{Inverse Problem }
\noindent A probability space is defined as a triplet $(\Omega,\ \mathcal{F}\ ,\ \mathbb{P}\mathrm{)}$. The sample space $\Omega$ is a set containing all possible outcomes. The $\sigma $-algebra $\mathcal{F}$ on $\Omega$ is a set of all possible events from an experiment, and follows the following axioms:
if $A\ \in \mathcal{F}$, then $A^c\in \mathcal{F}$, where $A^c$\textit{ }is the complement of $A$.  If $A_1,\ A_2,\ ...\ \in \ \mathcal{F}$ is a countable sequence of elements, then $U^{\infty }_{i=1}=A_i\ \in \ \mathcal{F}$ \cite{stuart2010inverse}
\noindent The measure $\mathbb{P}\ :\ \mathcal{F}\ \to \ \left[0,1\right]$ defines a probability to every event in $\mathcal{F}$ and satisfies:
$\mathbb{P}\left(A\right)\ge 0,\ \forall \ A\ \in \ \mathcal{F}$, $\mathbb{P}\left(\Omega\right)=1$, If $A_1,\ A_2,\ ...\ \in \ \mathcal{F}$ are pairwise disjoint (meaning $A_i\ \cap \ A_j=0,\ \forall \ i\ \neq j)$, then $\mathbb{P}\left(U^{\infty }_{i=1}=\left(A_i\right)\right)=\ \sum^{\infty }_{i=1}{\mathbb{P}(A_i)}$${}_{.\ }$This property is called countable additivity\cite{stuart2010inverse}.

If measurable sets $A_i$ for $i \in \mathbb{N}$ are pairwise disjoint, then $\mathbb{P}\left( \bigcup_{i=1}^{\infty} A_i \right) = \sum_{i=1}^{\infty} \mathbb{P}(A_i)$.
\noindent A measure is called $\sigma$-finite if $\Omega$ is a countable union of measurable sets with finite measure. In this paper, we restrict the Euclidean space $\Omega = \mathbb{R}^n$, with Borel $\sigma$-algebra $\mathcal{F} = \mathcal{B}(\mathbb{R}^n)$, and with Lebesgue measure on $\mathbb{R}^n$. A random variable $X$ is defined as a measurable map $X: (\Omega, \mathcal{F}) \to (\mathbb{R}^n, \mathcal{B}(\mathbb{R}^n))$, which means that if $A \in \mathcal{B}(\mathbb{R}^n)$, then $X^{-1}(A) \in \mathcal{F}$. The probability distribution of $X$ is then defined as a measure on $(\mathbb{R}^n, \mathcal{B}(\mathbb{R}^n))$: $\mu(A) := \mathbb{P}(X^{-1}(A))$.

\noindent \textbf{Theorem 1 }(Radon- Nikodym Theorem). \textit{Let }$\mu $\textit{ and }$\nu $\textit{ be two measures on the same measure space }$(\Omega,\ \mathcal{F})$\textit{. If }$\nu $\textit{ is }$\sigma $\textit{- finite and }$\mu \ \ll \ \nu $\textit{ then there exists a function }$f\ \in {\mathcal{L}}^1\ (\Omega,\ \mathcal{F}\ ,\ \nu )$\textit{ such that}
\begin{equation}\tag{7a}
\mu \left(A\right)=\int_A{f\left(x\right)d\nu \left(x\right)\ ,.\ \forall \ A\ \in \ \mathcal{F}}
\end{equation}
\noindent \textit{And }$f\left(x\right)=\ \frac{d\mu }{d\nu }(x)$\textit{ is called the Radon-Nikodym derivative of }$\mu $\textit{ with respect to }$\nu $\textit{}

Suppose $\mu$ is a probability measure on $(\mathbb{R}^n, \mathcal{B}(\mathbb{R}^n))$, $\mathcal{L}_n$ is a Lebesgue measure on $\mathbb{R}^n$, and $\mu \ll \mathcal{L}_n$, then there exists $\pi \in \mathcal{L}^1(\mathbb{R}^n)$ such that $\mu(A) = \int_A \pi(x) \, dx$ for any $A \in \mathcal{B}(\mathbb{R}^n)$, and $\pi(x)$ is called the probability density function (pdf) of $X$. The joint distribution of two random variables $X$ and $Y$ over corresponding measurable sets $A$ and $B$ is given by $\mu_{X \times Y}(A \times B) = \mathbb{P}(X^{-1}(A) \cap Y^{-1}(B))$, and the marginal distribution of $X$ is defined as $\mu_X(A) = \mu_{X \times Y}(A \times \mathbb{R}^n)$, similarly for $Y$. 
Two random variables $X$ and $Y$ are independent if $\mu_{X \times Y}(A \times B) = \mu_X(A) \mu_Y(B)$.

A random variable $Y \in \mathcal{L}^1(\Omega, \boldsymbol{G}, \mathbb{P}; \mathbb{R}^n)$ is called the conditional expectation of $X \in \mathcal{L}^1(\Omega, \mathcal{F}, \mathbb{P}; \mathbb{R}^n)$ with respect to the partial $\sigma$-algebra $\boldsymbol{G}$ if $\int_G X(\omega) \, d\mathbb{P}(\omega) = \int_G Y(\omega) \, d\mathbb{P}(\omega)$ for all $G \in \boldsymbol{G}$, then define the expectation $\mathbb{E}(X | \boldsymbol{G}) := Y$. The conditional probability of an event $A \in \mathcal{F}$ given the sub-$\sigma$-algebra $\boldsymbol{G}$ is a conditional expectation of the form $\mathbb{E}(f(x), \boldsymbol{G})$ where $\mathbb{P}(A | \boldsymbol{G}) = \mathbb{E}(\boldsymbol{1}_A(X) | \boldsymbol{G})$ and $\boldsymbol{1}_A(X) := \left\{ \begin{array}{c} 1 \text{ if } x \in A \\ 0 \text{ if } x \notin A \end{array} \right.$ is the indicator function of a subset $A$ of a set $X$.

\noindent The cost functional for the history matching inverse problem is posed as \cite{stuart2010inverse,emerick2013ensemble,etienam2023reservoir,chen2011ensemble,etienam2024novel,iglesias2013regularizing}.
\begin{equation}\tag{7b}
\mathrm{\Phi }\left(u;y\right)\equiv {\frac{1}{2}\left\|{\mathrm{\Gamma }}^{\frac{-1}{2}}\left(y-G\left(u\right)\right)\right\|}^2
\end{equation}
\noindent Where $\left\|.\right\|$ represents the $M$ dimensional Euclidean Norm. ~$G:\ {\mathbb{R}}^d\ \to \ {\mathbb{R}}^k$ represents the forward simulator, that solves the forward problem and defines the relationship between unknown parameters $u\ \in \ {\mathbb{R}}^d$ , $y\ \in \ {\mathbb{R}}^k\ $ is the corrupted noisy (or noiseless in rare cases) true data given by.
\begin{equation}\tag{7c}
y=F\left(K^+,{\phi }^+\right)+\ \eta 
\end{equation}
where $K^+,{\phi }^+$ is the underlining unknown true model (we want to recover) and $\eta \mathrm{\sim }\mathbb{N}\left(0,\ \mathrm{\Gamma }\right)$ . $\mathrm{\Gamma }\mathrm{\ }\in \ {\mathbb{R}}^{k\times k}$ is the data noise covariance matrix with Lebesgue density $\rho (\eta )$. The unknown $u$ follows a prior distribution ${\mu }_0$ with Lebesgue density ${\rho }_0(u)$. This prior density ${\rho }_0(u)$ contains any information that we already know about $u;$ a good prior is key to the solution of the inverse problem. It then follows that $\left(u,y\right)\in \ {\mathbb{R}}^d$ $\times {\mathbb{R}}^k\ $is a random variable with Lebesgue density $\rho \left(y-\ G\left(u\right)\ \right){\rho }_0(u)$.
\noindent The posterior distribution $u|y$ is then the solution to the inverse problem and is given by Bayes' theorem,
\begin{equation}\tag{7d}
\mathbb{P}\left(u\mathrel{\left|\vphantom{u y}\right.\kern-\nulldelimiterspace}y\right)=\ \frac{1}{\mathbb{P}\left(y\right)}\mathbb{P}\left(y\mathrel{\left|\vphantom{y u}\right.\kern-\nulldelimiterspace}u\right)\mathbb{P}(u)
\end{equation}
Assume that $Z := \int_{\mathbb{R}^n} \rho(y - G(u)) \rho_0(u) \, du > 0$, then $u|y$ is a random variable with Lebesgue density $\pi(u) = \frac{1}{Z} \rho(y - G(u)) \rho_0(u)$, where $Z$ is the probability of $y$ and acts as a normalizing constant; for random variables $A$ and $B$, let $\mathbb{P}(A|B)$ be the conditional pdf of $A$ given $B$ and $\mathbb{P}(A)$ be the pdf of $A$, then $\mathbb{P}(u, y) = \mathbb{P}(u|y) \mathbb{P}(y) = \mathbb{P}(y|u) \mathbb{P}(u)$, and $\mathbb{P}(y) = \int_{\mathbb{R}^d} \mathbb{P}(u, y) \, du$ and $\mathbb{P}(u) = \int_{\mathbb{R}^k} \mathbb{P}(u, y) \, dy$, assuming that $\mathbb{P}(y) > 0$ then Bayes' theorem gives us $\mathbb{P}(u|y) = \frac{1}{\mathbb{P}(y)} \mathbb{P}(y|u) \mathbb{P}(u) = \frac{1}{Z} \rho(y - G(u)) \rho_0(u) > 0$; if we define the negative log likelihood as $\Phi(u; y) = -\log \rho(y - G(u))$, then $\frac{d\mu^y}{d\mu_0}(u) = \frac{1}{Z} \exp(-\Phi(u; y))$, where $Z = \int_{\mathbb{R}^k} \exp(-\Phi(u; y)) \mu_0(du)$, and $\mu^y$ and $\mu_0$ are measures on $\mathbb{R}^k$ with densities $\pi$ and $\rho_0$ respectively, which shows the posterior is absolutely continuous with respect to the prior density and the Radon-Nikodym derivative $\frac{d\mu^y}{d\mu_0}(u)$ is proportional to the likelihood.
\noindent Using the Sherman -Morrison-Woodbury Lemma \cite{sherman1950adjustment}, we arrive at the Vanilla \textbf{E}nsemble \textbf{K}alman \textbf{I}nversion (EKI),
\begin{equation}\tag{8a}
u^{(j)}_{n+1}=u^{(j)}_n+C^{uG}_n{\left(C^{GG}_n+{\alpha }_n\mathrm{\Gamma }\right)}^{-1}(y\ +\ \sqrt{\alpha }{\xi }_n-G(u^{(j)}_n))
\end{equation}
\noindent $n$ is the iteration index, $j$ is the ensemble member (columns in the matrix) in the ensemble size $J$. ${\alpha }_n$ is a (crucial) regularisation dampener, ~$C^{uG}_n$ and $C^{GG}_n$ are cross-covariance matrixes to be explained shortly, $\xi \mathrm{\ }\mathrm{\sim }\mathbb{N}\mathrm{(}0,1)$.
\begin{equation}\tag{8b}
C^{GG}_n=\ \frac{1}{j-1}\sum^J_{j=1}{\left(G\left(u^{\left(j\right)}_n\right)-\overline{G\left(u^{\left(j\right)}_n\right)}\right)\otimes }\left(G\left(u^{\left(j\right)}_n\right)-\overline{G\left(u^{\left(j\right)}_n\right)}\right)
\end{equation}
\begin{equation}\tag{8c}
C^{uG}_n=\ \frac{1}{j-1}\sum^J_{j=1}{\left(\left(u^{\left(j\right)}_n\right)-\overline{\left(u^{\left(j\right)}_n\right)}\right)\otimes }\left(G\left(u^{\left(j\right)}_n\right)-\overline{G\left(u^{\left(j\right)}_n\right)}\right)
\end{equation}
\noindent In the developed \textit{a}REKI method, the dampening parameter $\alpha $ is developed via the \textit{discrepancy principle} \cite{tarantola2005inverse,stuart2010inverse} in Algorithm 2.
\begin{algorithm}
\small
\caption{\textit{a}REKI}
\small
\label{alg2}
\begin{algorithmic}[1]
\Require
\begin{itemize}
    \item ${\left\{u^{(j)}_0\right\}}^J_{j=1}$ \Comment{prior density}
    \item $y$ \Comment{measurements}
    \item $\mathrm{\Gamma }$ \Comment{measurement errors covariance}
    \item $iter$ \Comment{iteration}
\end{itemize}
\State Initialize: $s_n = 0$, $n = 0$
\While{$(s_n < 1)$ or $(n \leq iter)$}
    \State Compute: $G\left(u^{(j)}_n\right),\ j \in \{1,\ldots,J\}$
    \State Compute: ${\mathrm{\Phi }}_n\equiv {\left\{\frac{1}{2}{\left\|{\mathrm{\Gamma }}^{\frac{-1}{2}}\left(y-G\left(u\right)\right)\right\|}^2_2\right\}}^J_{j=1}$
    \State Compute: $\overline{{\mathrm{\Phi }}_n}$, ${\sigma }^2_{{\mathrm{\Phi }}_n}$ \Comment{mean \& variance}
    \State Compute: ${\alpha }_n = \frac{1}{\mathrm{min} \left\{\mathrm{max} \left\{\frac{L_{(y)}}{2\overline{{\mathrm{\Phi }}_n}},\ \sqrt{\frac{L_{(y)}}{2\overline{{\sigma }^2_{{\mathrm{\Phi }}_n}}}}\right\}, 1 - s_n \right\}}$
    \State Compute: $s_n = s_n + \frac{1}{{\alpha }_n}$
    \State Compute: $C^{GG}_n= \frac{1}{j-1}\sum^J_{j=1}{\left(G\left(u^{(j)}_n\right)-\overline{G\left(u^{(j)}_n\right)}\right)\otimes }\left(G\left(u^{(j)}_n\right)-\overline{G\left(u^{(j)}_n\right)}\right)$
    \State Compute: $C^{uG}_n= \frac{1}{j-1}\sum^J_{j=1}{\left(\left(u^{(j)}_n\right)-\overline{\left(u^{(j)}_n\right)}\right)\otimes }\left(G\left(u^{(j)}_n\right)-\overline{G\left(u^{(j)}_n\right)}\right)$
    \State Update ensemble: $u^{(j)}_{n+1} = u^{(j)}_n + C^{uG}_n{\left(C^{GG}_n + {\alpha }_n\mathrm{\Gamma }\right)}^{-1}\left(y + \sqrt{{\alpha }_n}{\xi }_n - G\left(u^{(j)}_n\right)\right)$
    \State Increment: $n \leftarrow n + 1$
\EndWhile
 \State \textbf{Output:}: $u^{(j)}$\Comment{posterior density}
\end{algorithmic}
\end{algorithm}
Covariance localization means localising the effect of an observation to the state variables that are `closer' to the observations. It is a technique used in history matching to eliminate filter divergence and spurious correlations which are found in both standard ensemble-based methods \cite{evensen2003ensemble,chen2011ensemble,emerick2013ensemble,oliver2010recent,oliver2008inverse}. It selectively assimilates data during the update step by adjusting the correlation between the data measured at a location A and the model parameters or data measured at a further location B to be zero, therefore eliminating long-distance non-zero spurious correlations and increasing the degrees of freedom available for data assimilation. It is usually applied with a Schur product.
\noindent The various localization methods proposed in the literature have the common goal of removing the spurious terms in the cross-covariance matrix; this matrix is in turn used to update the state vectors during the ensemble update process. This is done by conditioning the Kalman Gain through a localizing function \cite{evensen2003ensemble,chen2011ensemble,emerick2013ensemble,oliver2010recent,oliver2008inverse}. Each localization scheme distinguishes itself from each other by how this localizing function, also known as the Schur product or the multiplier function, is computed. The motivation of covariance localization is to achieve a similar level \textit{a}REKI performance if a larger ensemble size would have been used. However, it requires enormous computational resources to perform history matching if the ensemble size is large.
\noindent $\rho $ is the localization function or multiplier function, this is a specified parameter for covariance localization. It is introduced to modify the Kalman Gain during update step for \textit{a}REKI. The symbol `◦' is an element-by-element multiplication operator known as the Schur product.
\noindent To implement the Schur product in covariance localization, a correlation function with local support, $\rhoup$ is defined. The term local support refers to the function taking on non-zero values in a small region and being zero elsewhere. The function $\rhoup$ is usually defined to be the compactly supported fifth order piecewise rational function defined by Gaspari and Cohn as shown below \cite{evensen2003ensemble,chen2011ensemble,emerick2013ensemble,oliver2010recent,oliver2008inverse}.
\begin{equation}\tag{9a}
\rho \mathrm{=}\left\{ \begin{array}{cc}
\mathrm{-}\frac{\mathrm{1}}{\mathrm{4}}{\left(\frac{\left|z\right|}{c}\right)}^{\mathrm{5}}\mathrm{+}\frac{\mathrm{1}}{\mathrm{2}}{\left(\frac{\left|z\right|}{c}\right)}^{\mathrm{4}}\mathrm{+}\frac{\mathrm{5}}{\mathrm{8}}{\left(\frac{\left|z\right|}{c}\right)}^{\mathrm{3}}\mathrm{-}\frac{\mathrm{5}}{\mathrm{3}}{\left(\frac{\left|z\right|}{c}\right)}^{\mathrm{2}}\mathrm{+1,} & ~\mathrm{0}\mathrm{\le }\left|z\right|\mathrm{\le }c, \\ 
\frac{\mathrm{1}}{\mathrm{12}}{\left(\frac{\left|z\right|}{c}\right)}^{\mathrm{5}}\mathrm{-}\frac{\mathrm{1}}{\mathrm{2}}{\left(\frac{\left|z\right|}{c}\right)}^{\mathrm{4}}\mathrm{+}\frac{\mathrm{5}}{\mathrm{8}}{\left(\frac{\left|z\right|}{c}\right)}^{\mathrm{3}}\mathrm{+}\frac{\mathrm{5}}{\mathrm{3}}{\left(\frac{\left|z\right|}{c}\right)}^{\mathrm{2}}\mathrm{-}\mathrm{5}\left(\frac{\left|z\right|}{c}\right)\mathrm{+4-}\frac{\mathrm{2}}{\mathrm{3}}\left(\frac{\left|z\right|}{c}\right), & c\mathrm{\le }\left|z\right|\mathrm{\le }\mathrm{2}c, \\ 
0, & ~~\mathrm{2}c\mathrm{\le }\left|z\right|. \end{array}
\right.
\end{equation}
\begin{equation}\tag{9b}
u^{(j)}_{n+1} = u^{(j)}_n + \rho \cdot o \cdot C^{uG}_n \left(C^{GG}_n + \alpha_n \Gamma\right)^{-1} \left(y + \sqrt{\alpha_n} \xi_n - G\left(u^{(j)}_n\right)\right), \quad j \in \{1, \ldots, J\}
\end{equation}
\section{Mixture of experts. }
In general, the problem of designing machine learning-based models in a supervised context is the following. Assume a set of labeled data $\boldsymbol{D} = \{(x_i, y_i)\}_{i=1}^N$, where $x_i \in \mathbb{R}^d$ are the inputs and the outputs $y_i \in \mathbb{R}$ for regression or $y_i \in \{0, 1, \ldots, J\}$ for classification. Postulating amongst a family of ansatz $f(\cdot; \theta)$, parametrized by $\theta \in \mathbb{R}^P$, we can find a $\theta^*$ such that for all $i = 1, \ldots, N$, $y_i = f(x_i; \theta^*)$. Then for all $x'$ in the set $\{x_i\}_{i=1}^N$, $y' \approx f(x'; \theta^*)$, where $y'$ is the true label of $x'$. \cite{etienam2023ultrafast,trapp2020deep,etienam2024novel} . Assume $\left(x_i,y_i\right)\in \ \chi \ \times \ \gamma $ are the input and output pairs of a model shown in Eqn.  \ref{10a}.
\begin{equation}\tag{10a}
y_i\approx f\left(x_i\right)
\label{10a}
\end{equation}
\noindent $f:\chi \to \gamma $ is irregular and has sharp features, very non-linear and has noticeable discontinuities, The output space is taken as  $\gamma \mathrm{=}\mathbb{R}$ and $\chi ={\mathbb{R}}^d$
\subsubsection{\textbf{Cluster}}
\noindent In this stage, we seek to cluster the training input and output pairs.
\noindent $\lambda :\chi \ \times \ \gamma \ \to \mathcal{L}\coloneqq \left\{1,\dots ,L\right\}$ where the label function minimizes,
\begin{equation}\tag{10b}
{\mathrm{\Phi }}_{clust}\left(\lambda \right)=\sum^L_{l=1}{\sum_{i\epsilon S_l}{\ell \left(x_i,y_i\right)}}
\end{equation}
\begin{equation}\tag{10c}
S_l=\left\{\left(x_i,y_i\right);\lambda \left(x_i,y_i\right)=l\right\}
\end{equation}
\noindent ${\ell }_l$ loss function associated to cluster $l$
\begin{equation}\tag{10d}
z_i=\left(x_i,y_i\right),{\ell }_l={\left|z_i-{\mu }_l\right|}^2
\end{equation}
\noindent ${\mu }_l=\frac{1}{\left|S_l\right|}\sum_{i\epsilon S_l\ }{z_i}$ where $\left|.\right|$ denotes the Euclidean norm
\subsubsection{\textbf{Classify}}
\noindent $l_i=\lambda \left(x_i,y_i\right)$ is an expanded training set
\begin{equation}\tag{11a}
{\left\{\left(x_i,y_i,l_i\right)\right\}}^N_{i=1}
\end{equation}
\noindent 
\begin{equation}\tag{11b}
f_c:\chi \to \mathcal{L}
\end{equation}
\noindent $x\in \chi \ \ $provides an estimate $f_c\ :x\longmapsto f\left(x\right)\in \ \mathcal{L}\ $such that $f_c\left(x_i\right)=l_i\ $for most of the data. Crucial for the ultimate fidelity of the prediction. $\left\{y_i\right\}$ is ignored at this phase. The classification function minimizes,
\begin{equation}\tag{11c}
{\mathrm{\Phi }}_{clust}\left(f_c\right)=\sum^N_{i=1}{{\phi }_c\left(l_i,f_c\left(x_i\right)\right)}
\end{equation}
\noindent ${\phi }_c\ :\ \mathcal{L}\times \ \mathcal{L}\ \to \ {\mathbb{R}}_+\ $ is small if $f_c\left(x_i\right)=l_i$ for example we can choose $f_c\left(x\right)={argmax}_{l\in \mathcal{L}}g_l\left(x\right)$ where $g_l\left(x\right)>0,\ \sum^{\mathcal{L}}_{l=1}{g_l\left(x\right)}$ is a soft classifier and ${\phi }_c\left(l_i,f_c\left(x_i\right)\right)=-{\mathrm{log} \left(g_l\left(x\right)\right)\ }$ is a cross-entropic loss.
\subsubsection{\textbf{Regress}}
\begin{equation}\tag{11d}
f_r\ :\chi \ \times \mathcal{L}\ \to \ \gamma \ \
\end{equation}
\noindent For each $\left(x,l\right)$ $\in \ \chi \ \times \mathcal{L}\ $must provide an estimate $f_r\ :\left(x,l\right)\ \longmapsto \ f_r\left(x,l\right)\ \in \ \ \gamma \ $such that $f_r\left(x,\ f_c\left(x\right)\right)\ \approx y$ for both the training and test data. If successful a good reconstruction for 
\begin{equation}\tag{11e}
f:\chi \to \gamma
\end{equation}
\noindent Where $f\left(.\right)=f_r\left(.,\ f_c\left(.\right)\right)$ the regression function can be found by minimizing
\begin{equation}\tag{11f}
{\mathrm{\Phi }}_r\left(f_r\right)=\sum^N_{i=1}{{\phi }_r\left(y_i{,f}_r\left(x_i,\ f_c\left(x_i\right)\right)\right)}
\end{equation}
\noindent Where ${\phi }_r\ :\ \gamma \times \ \gamma \ \to \ {\mathbb{R}}_+$ minimized when $f_r\left(x_i,\ f_c\left(x_i\right)\right)=y_i$ in this case can be chosen as ${\phi }_r\left(y{,f}_r\left(x,\ f_c\left(x\right)\right)\right)=\ {\left|{y-f}_r\left(x,\ f_c\left(x\right)\right)\right|}^2$. Data can be partitioned into $C_l=\left\{i;f_c\left(x_i\right)=l\right\}$ for $l=1,\dots L$ and then perform $L$ separate regressions done in parallel.
\begin{equation}\tag{11g}
{\mathrm{\Phi }}^l_r(f_r\left(.,l\right)=\sum_{i\in C_l}{{\phi }_r\left(y_i{,f}_r\left(x_i,\ l\right)\right)}
\end{equation}
\subsubsection{\textbf{Bayesian Formulation}}
\noindent A critique of this method is that it re-uses the data in each phase. A Bayesian postulation handles this limitation elegantly. Recall,
\begin{equation}\tag{12a}
{\boldsymbol{D}=\left\{\left(x_i,y_i\right)\right\}}^N_{i=1}
\end{equation}
\noindent Assume parametric models for the classifier $g_l\left(.;{\theta }_c\right)=g_l\left(.;{\theta }^l_c\right)$ and the regressor. 
\noindent $f_r\left(.,l;{\theta }^l_r\right)$ for $l=1,\dots L$ where ${\theta }_c=\left({\theta }^1_c,\dots ,{\theta }^L_c\right)$ and ${\theta }_r=\left({\theta }^1_r,\dots ,{\theta }^L_r\right)$ and let $\theta =\left({\theta }_c,{\theta }_r\right)$, the posterior density has the form,
\begin{equation}\tag{12b}
\pi \left(\theta ,l\mathrel{\left|\vphantom{\theta ,l D}\right.\kern-\nulldelimiterspace}D\right)\propto \prod^N_{i=1}{\pi \left(y_i\mathrel{\left|\vphantom{y_i x_i,{\theta }_r,l}\right.\kern-\nulldelimiterspace}x_i,{\theta }_r,l\right)}\pi \left(l\mathrel{\left|\vphantom{l x_i,{\theta }_c}\right.\kern-\nulldelimiterspace}x_i,{\theta }_c\right)\pi \left({\theta }_r\right)\pi \left({\theta }_c\right)
\end{equation}
\begin{equation}\tag{12c}
\pi \left(y_i\mathrel{\left|\vphantom{y_i x_i,{\theta }_r,l}\right.\kern-\nulldelimiterspace}x_i,{\theta }_r,l\right)\propto {\mathrm{exp} \left(-\frac{1}{2}{\left|y_i-f_r\left(x_i,l;{\theta }^l_r\right)\right|}^2\right)\ }
\end{equation}
\noindent And 
\begin{equation}\tag{12d}
\pi \left(l\mathrel{\left|\vphantom{l x_i,{\theta }_c}\right.\kern-\nulldelimiterspace}x_i,{\theta }_c\right)=g_l\left(x_i;{\theta }_c\right)
\end{equation}
\begin{equation}\tag{12e}
g_l\left(x_i;{\theta }_c\right)=\frac{{\mathrm{exp} \left(h_l\left(x;{\theta }^l_c\right)\right)\ }}{\sum^L_{l=1}{{\mathrm{exp} \left(h_l\left(x;{\theta }^l_c\right)\right)\ }}}
\end{equation}
\noindent $h_l\left(x;{\theta }^l_c\right)$ are some standard parametric regressors.
\subsection{Sparse Gaussian process experts}
Mixtures of GP experts have proven to be very successful 
\cite{snelson2006, titsias2009, bui2016, etienam2023ultrafast, CarlRass2010}. In particular, they overcome limitations of stationary Gaussian process models by reducing the computational complexity through local approximations and allow different local properties of the unknown function to handle challenges, such as discontinuities, non-stationarity and non-normality. In this case, one assumes a GP prior on the local regression function with hyperparameters 
$ \theta_r^l =( \mu^l,\psi^l)$:
 $$ f_r(\cdot; \theta_r^l) \sim \text{GP}(\mu^l, K_{\psi^l}),$$
 where $\mu^l$ is the local mean function of the expert (for simplicity, it assumed to be constant) and $\psi^l$ are the parameters of the covariance function $K_{\psi^l}$, whose chosen form and hyperparameters encapsulate properties of the local function such as the spatial correlation, smoothness, and periodicity. 
While GP experts are appealing due to their flexibility, intrepretability and probabilistic nature, they increase the computational cost of the model significantly. Indeed, given the allocation variables $\bz$,
the GP hyperparameters, which crucially determine the behavior of the unknown function, can be estimated by optimizing the log marginal likelihood:
\begin{equation}\tag{12f}
\log(p(\by | \bx, \bz)) = \sum_{l=1}^L  \log\left( \text{N}(\by^l \mid \bm{\mu}^l, \mathbf{K}^l_{N_l}+\sigma^{2 \, l}_r \mathbf{I}_{N_l} )\right)\, ,
\end{equation}
where $\by^l$ and $\bx^l$ contain the outputs and inputs of the $l^\text{th}$ cluster, i.e. $\by^l = \lbrace y_i \rbrace_{z_i=l}$ and $\bx^l = \lbrace x_i \rbrace_{z_i=l}$; $\bm{\mu}^l $ is a vector with entries $\mu^l$; $\mathbf{K}^l_{N_l}$ represents the $N_l \times N_l$ matrix obtained by evaluating the covariance function $K_{\psi^l}$ at each pair of inputs in the $l^\text{th}$ cluster; and $N_l$ is number data points in the $l^\text{th}$ cluster. This however requires inversion of $N_l \times N_l$ matrices, which scales $\cO(\sum_{l=1}^L N_l^3)$. While this reduces the computational complexity compared with standard GP models which scale $\cO(N^3)$, it can still be costly\cite{etienam2023ultrafast}.

To improve scalability, one can resort to approximate methods for GPs, including sparse GPs based on a set of inducing points or pseudo inputs \cite{snelson2006, titsias2009, bui2016,etienam2023ultrafast}, basis function approximations \cite{bui2016}, or sparse formulations of the precision matrix \cite{titsias2009, bui2016}, among others  for a recent review of approaches in spatial statistics and \cite{titsias2009, bui2016} for a review of approaches in machine learning).  

In the present work, we employ an inducing point strategy, 
assuming the local likelihood of the data points within each cluster factorizes given a set of $M_l<N_l$ pseudo-inputs $\tilde{\bx}^l = (\tilde{x}^l_1, \ldots, \tilde{x}^l_{M_l})$ 
and pseudo-targets $\tilde{\bof}^l= (\tilde{f}^l_1, \ldots, \tilde{f}^l_{M_l}) $; 
\begin{equation}\tag{12g} p(\by^l | \bx^l, \tilde{\bx}^l , \tilde{\bof}^l) = \prod_{i:z_i=l} \text{N}(y_i| \widehat{\mu}^l_i, 
 \widehat{\sigma}^{2 \, l}_i), \label{eq:sparse_like}
 \end{equation}
where
$$\widehat{\mu}^l_i= 
\mu^l + (\mathbf{k}^l_{M_l,i})^T(\mathbf{K}^l_{M_l})^{-1} (\tilde{\bof}^l- \bm{\mu}^l),$$
 $$\widehat{\sigma}^{2 \, l}_i = 
\sigma^{2 \, l}_r + K_{\psi^l}(x_i,x_i)- 
(\mathbf{k}^l_{M_l,i})^T(\mathbf{K}^l_{M_l})^{-1}\mathbf{k}^l_{M_l,i},$$
where $\mathbf{K}^l_{M_l}$ is the $M_l \times M_l$ matrix with elements $K_{\psi^l}(\tilde{x}^l_j,\tilde{x}^l_h)$ and $\mathbf{k}^l_{M_l,i}$ is the vector of length $M_l$ with elements $K_{\psi^l}(\tilde{x}^l_j, x_i)$. 
This corresponds to the fully independent training conditional (FITC) approximation \cite{quinonero2005}. After marginalization of the pseudo-targets  under the GP prior $\tilde{\bof}^l \sim \text{N}(\bm{\mu}^l, \mathbf{K}^l_{M_l})$, the pseudo-inputs $\tilde{\bx}^l $ and hyperparameters $(\mu^l, \psi^l)$ can be estimated by optimizing the marginal likelihood:
\begin{equation}\tag{12h}
\log(p(\mathbf{y} | \mathbf{x}, \mathbf{z}, \tilde{\mathbf{x}})) = \sum_{l=1}^L \log\left( \mathcal{N}(\mathbf{y}^l \mid \bm{\mu}^l, (\mathbf{K}^l_{M_l N_l})^T (\mathbf{K}^l_{M_l})^{-1} \mathbf{K}^l_{M_l N_l} + \bm{\Lambda}^l + \sigma^{2 \, l}_r \mathbf{I}_{N_l}) \right) ,
\label{eq:sparse_marg}
\end{equation}

where $\mathbf{K}^l_{M_l N_l}$ is the $M_l \times N_l$ matrix with columns $\mathbf{k}^l_{M_l,i}$ and $\bm{\Lambda}^l$ is the diagonal matrix with diagonal entries $K_{\psi^l}(x_i,x_i)- 
(\mathbf{k}^l_{M_l,i})^T(\mathbf{K}^l_{M_l})^{-1}\mathbf{k}^l_{M_l,i}$. This strategy allows us to reduce the complexity to $\cO(\sum_{l=1}^L N_l M_l^2)$.
\section{Generative priors and Tree based learning}
\subsection{Variational Convolutional Autoencoder (VCAE)}
A Variational Convolutional Autoencoder (VCAE) is an extension of the Variational Autoencoder (VAE) with convolutional layers to handle image data more effectively\cite{kingma2014autoencoding}.
\subsubsection{Encoder}\label{encoderr}
The encoder is a convolutional neural network that maps the input data \( x \in \mathbb{R}^d \) to the parameters of a variational distribution over the latent variables \( z \in \mathbb{R}^k \). The encoder outputs the mean and log variance of a Gaussian distribution:
\begin{equation}\tag{13a}
q_\phi(z|x) = \mathcal{N}(z; \mu_\phi(x), \sigma_\phi^2(x) \mathbf{I})
\end{equation}
\begin{equation}\tag{13b}
\mu_\phi(x) = \text{ConvNet}_\mu(x)
\end{equation}
\begin{equation}\tag{13c}
\log \sigma_\phi^2(x) = \text{ConvNet}_{\log \sigma^2}(x)
\end{equation}
where \(\phi\) denotes the parameters of the encoder network.
\subsubsection{Decoder}
The decoder is a deconvolutional neural network that maps the latent variables \( z \) back to the data space, producing the parameters of a generative distribution over \( x \):
\begin{equation}\tag{13d}
p_\theta(x|z) = \mathcal{N}(x; \mu_\theta(z), \sigma_\theta^2(z) \mathbf{I})
\end{equation}
where \(\theta\) denotes the parameters of the decoder network, and \(\mu_\theta(z)\) and \(\log \sigma_\theta^2(z)\) are outputs of the deconvolutional network.
\subsubsection{Evidence Lower Bound (ELBO)}
The VCAE maximizes the evidence lower bound (ELBO) to approximate the true posterior distribution \( p(z|x) \):
\begin{equation}\tag{14a}
\mathcal{L}(\theta, \phi; x) = \mathbb{E}_{q_\phi(z|x)}[\log p_\theta(x|z)] - D_{\text{KL}}(q_\phi(z|x) \| p(z))
\end{equation}
\begin{equation}\tag{14b}
= \mathbb{E}_{q_\phi(z|x)}[\log \mathcal{N}(x; \mu_\theta(z), \sigma_\theta^2(z) \mathbf{I})] - D_{\text{KL}}(\mathcal{N}(z; \mu_\phi(x), \sigma_\phi^2(x) \mathbf{I}) \| \mathcal{N}(0, \mathbf{I}))
\end{equation}
The first term is the reconstruction loss, and the second term is the Kullback-Leibler (KL) divergence between the approximate posterior \( q_\phi(z|x) \) and the prior \( p(z) \).
\begin{algorithm}[H]
\small  
\caption{CCR}
\small
\label{alg3}
\begin{algorithmic}[1]
\State \textbf{Input:} Set of labeled data $\boldsymbol{D} = \{(x_i, y_i)\}_{i=1}^N$ where $x_i \in \mathbb{R}^d$ and $y_i \in \mathbb{R}$
\State \textbf{Output:} Refined model estimation $f(x; \theta_{\text{ccr}}) = f_r(x, f_c(x))$
\State
\Function{Cluster}{$\boldsymbol{D}$}
    \State Initialize $\lambda : \mathbb{R}^d \times \mathbb{R} \to \{1, \ldots, L\}$
    \For{each $(x_i, y_i) \in \boldsymbol{D}$}
        \State Assign $(x_i, y_i)$ to a cluster $l = \lambda(x_i, y_i)$
        \State Calculate $S_l = \{(x_i, y_i) : \lambda(x_i, y_i) = l\}$
        \State Calculate cluster center $\mu_l = \frac{1}{|S_l|} \sum_{i \in S_l} (x_i, y_i)$
    \EndFor
    \State Minimize $\Phi_{clust}(\lambda) = \sum_{l=1}^L \sum_{i \in S_l} |z_i - \mu_l|^2$
\EndFunction
\State
\Function{Classify}{$\boldsymbol{D}, \lambda$}
    \For{each $(x_i, y_i) \in \boldsymbol{D}$}
        \State Classify $l_i = \lambda(x_i, y_i)$
        \State Define $f_c(x_i) \gets \operatorname{argmax}_{l \in \mathcal{L}} g_l(x_i)$
    \EndFor
    \State Minimize $\Phi_{clust}(f_c) = \sum_{i=1}^N -\log(g_{l_i}(x_i))$
\EndFunction
\State
\Function{Regress}{$\boldsymbol{D}, f_c$}
    \For{each $(x_i, y_i, l_i)$}
        \State Estimate $f_r(x_i, l_i) \approx y_i$ using the regression model
        \State Minimize $\Phi_r(f_r) = \sum_{i=1}^N (y_i - f_r(x_i, f_c(x_i)))^2$
    \EndFor
\EndFunction
\end{algorithmic}
\end{algorithm}
\begin{algorithm}[H]
\caption{Variational Convolutional Autoencoder (VCAE)}
\label{alg4}
\begin{algorithmic}[1]

\State Initialize encoder parameters $\phi$ and decoder parameters $\psi$
\State Input image $x$
\State Encode $x$ to get latent variables: $z, \mu, \sigma = \text{Encoder}_\phi(x)$
\State Sample latent space: $z \sim \mathcal{N}(\mu, \sigma^2)$
\State Decode $z$ to reconstruct the image: $\hat{x} = \text{Decoder}_\psi(z)$
\State Compute loss: $\mathcal{L} = \text{Reconstruction Loss}(\hat{x}, x) + \text{KL Divergence}(\mu, \sigma)$
\State Update parameters $\phi$ and $\psi$ to minimize $\mathcal{L}$
 \State \textbf{Output}: Return the reconstructed image $\hat{x}$
\end{algorithmic}
\end{algorithm}
\subsection{Denoising Diffusion Implicit Models (DDIM)}
Denoising Diffusion Implicit Models (DDIMs) are a type of generative model that produce samples by reversing a forward diffusion process.\cite{song2020denoising}
\subsubsection{Forward Diffusion Process}
The forward diffusion process incrementally adds Gaussian noise to the data \( x_0 \) through a sequence of latent variables \( x_t \):
\begin{equation}\tag{15a}
q(x_t | x_{t-1}) = \mathcal{N}(x_t; \sqrt{\alpha_t} x_{t-1}, (1 - \alpha_t) \mathbf{I})
\end{equation}
where \(\alpha_t\) are the variance preserving coefficients that determine the noise level at each diffusion step.
\subsubsection{Reverse Diffusion Process}
In DDIMs, the reverse diffusion process aims to reconstruct the original data by gradually denoising the sequence of noisy samples:
\begin{equation}\tag{15b}
p(x_{t-1} | x_t) = \mathcal{N}(x_{t-1}; \mu_\theta(x_t, t), \sigma_\theta^2(x_t, t) \mathbf{I})
\end{equation}
where \(\mu_\theta(x_t, t)\) and \(\sigma_\theta^2(x_t, t)\) are parametrized by a neural network, typically conditioned on \(t\) to capture the step-dependent dynamics.
\subsubsection{Sampling}
Sampling from a DDIM involves iteratively applying the reverse diffusion process starting from noise:
\begin{equation}\tag{16a}
x_T \sim \mathcal{N}(0, \mathbf{I})
\end{equation}
\begin{equation}\tag{16b}
x_{t-1} = \mu_\theta(x_t, t) + \sigma_\theta(x_t, t) \epsilon, \quad \epsilon \sim \mathcal{N}(0, \mathbf{I})
\end{equation}
The sequence continues until \( x_0 \) is reached, effectively denoising the data to retrieve an approximation of the original sample.
\begin{algorithm}
\small
\caption{Denoising Diffusion Implicit Model (DDIM)}
\label{alg5}
\begin{algorithmic}[1]
\State Initialize parameters $\theta$
\State Set number of diffusion steps $T$
\State Sample initial noise $x_T \sim \mathcal{N}(0, I)$
\For{$t = T$ to $1$}
    \State Compute deterministic step $\hat{x}_{t-1} = f_\theta(x_t, t)$, where $f$ is the denoising function
\EndFor
 \State \textbf{Output:}: Return the generated sample $\hat{x}_0$
\end{algorithmic}
\end{algorithm}
\subsection{XGBoost}
The objective function in XGBoost can be expressed as \cite{chen2016xgboost}}:
\begin{equation}\tag{16c}
\mathcal{L}(\Theta) = \sum_{i=1}^{n} l(\hat{y}_i, y_i) + \sum_{k=1}^{K} \Omega(f_k)
\end{equation}
where:
\begin{itemize}
    \item \( \hat{y}_i \) is the predicted value for the $i$-th instance.
    \item \( y_i \) is the true value for the $i$-th instance.
    \item \( l \) is a differentiable convex loss function that measures the difference between the prediction \( \hat{y}_i \) and the target \( y_i \).
    \item \( f_k \) represents the $k$-th tree in the model.
    \item \( \Omega(f_k) \) is the regularization term which helps to smooth the final learnt weights to avoid overfitting.
\end{itemize}
\subsubsection{Additive Training}
XGBoost builds the model in an additive manner:
\begin{equation}\tag{17a}
\hat{y}_i^{(t)} = \hat{y}_i^{(t-1)} + f_t(x_i)
\end{equation}
where:
\begin{itemize}
    \item \( \hat{y}_i^{(t)} \) is the prediction of the $i$-th instance at the $t$-th iteration.
    \item \( f_t \) is the function (tree) added in the $t$-th iteration.
\end{itemize}
\subsubsection{Model Complexity}
The regularization term \( \Omega \) is defined as:
\begin{equation}\tag{17b}
\Omega(f) = \gamma T + \frac{1}{2} \lambda \sum_{j=1}^{T} w_j^2
\end{equation}
where:
\begin{itemize}
    \item \( T \) is the number of leaves in the tree.
    \item \( w_j \) is the score on the $j$-th leaf.
    \item \( \gamma \) and \( \lambda \) are regularization parameters.
\end{itemize}
\subsubsection{Approximate Tree Learning}
To learn the tree structure, XGBoost uses a greedy algorithm. For a given tree structure, the objective function can be written as:
\begin{equation}\tag{17c}
\mathcal{L}^{(t)} \approx \sum_{i=1}^{n} \left[ l(y_i, \hat{y}_i^{(t-1)} + f_t(x_i)) \right] + \Omega(f_t)
\end{equation}
Using a second-order Taylor expansion, we can approximate the loss function:
\begin{equation}\tag{17d}
\mathcal{L}^{(t)} \approx \sum_{i=1}^{n} \left[ g_i f_t(x_i) + \frac{1}{2} h_i f_t(x_i)^2 \right] + \Omega(f_t)
\end{equation}
where:
\begin{itemize}
    \item \( g_i = \frac{\partial l(y_i, \hat{y}_i)}{\partial \hat{y}_i} \) is the first derivative of the loss function.
    \item \( h_i = \frac{\partial^2 l(y_i, \hat{y}_i)}{\partial \hat{y}_i^2} \) is the second derivative of the loss function.
\end{itemize}
The structure score of a tree can then be calculated as:
\begin{equation}\tag{17e}
\mathcal{L}_{\text{split}} = \frac{1}{2} \left[ \frac{(\sum_{i \in I_L} g_i)^2}{\sum_{i \in I_L} h_i + \lambda} + \frac{(\sum_{i \in I_R} g_i)^2}{\sum_{i \in I_R} h_i + \lambda} - \frac{(\sum_{i \in I} g_i)^2}{\sum_{i \in I} h_i + \lambda} \right] - \gamma
\end{equation}
where \( I \) is the set of indices of data points, \( I_L \) and \( I_R \) are the sets of indices for the left and right children after the split, respectively.
\subsubsection{Prediction}
The final prediction for a given instance \( x_i \) is:
\begin{equation}\tag{17f}
\hat{y}_i = \sum_{t=1}^{T} f_t(x_i)
\end{equation}
where \( T \) is the total number of trees.
\section{Methodology}
We use NVIDIA Modulus \cite{Modulus}, an open-source deep-learning framework for building, training, and fine-tuning deep learning models using state-of-the-art physics ML methods to learn the PINO surrogate. This framework is part of NVIDIA's suite of AI tools and is tailored for domain scientists, engineers, and AI researchers working on complex simulations across various domains such as fluid dynamics, atmospheric sciences, structural mechanics, electromagnetics, acoustics, and related domains. The emphasis on providing components and models that introduce inductive bias into the neural networks (e.g. FNOs, GNNs, etc.) and training (e.g. physics-based constraints) allows for more accurate and generalizable trained models for physics-ML applications, especially in scenarios where the data might be scarce or expensive to obtain.  

The Modulus framework is fundamentally composed of Modulus Core (generally referred to as Modulus) and Modulus-Sym. Modulus-Sym is built on top of Modulus which is a generalized toolkit for physics-ML model development and training. Modulus provides core utilities to train AI models in the physics-ML domain with optimized implementations of various network architectures specifically suited for physics.based applications (MLPs, operator networks like FNOs, DeepONet, graph networks), efficient data pipes for typical physics-ML datasets (structured and unstructured grids, point-clouds, etc.) and utilities to distribute and scale the training across multiple GPUs and nodes. Modulus-Sym provides utilities to introduce physics-based constraints into the model training and has utilities to handle various geometries (via built-in geometry module or STLs), introduce PDE constraints, and compute gradients efficiently through an optimized gradient backend. It also provides abstractions such as pre-defined training loops that enable domain scientists to experiment with AI model training quickly without diving into the depths of specific implementations of DL training. Modulus-Sym is designed to be highly flexible, supporting various types of partial differential equations (PDEs) (built-in PDE definitions as well as easy ways to introduce your own PDEs symbolically) and constraint definitions typically found in physics-based problems like boundary and equation residual conditions.  Modulus is compatible with NVIDIA's broader AI and HPC (high-performance computing) ecosystems. It is built on top of PyTorch and is interoperable with PyTorch. Modulus’ architecture allows for easy customization and has several integrations with other NVIDIA technologies like CUDA, Warp, PySDF, DALI, NVFuser, and Omniverse, to facilitate efficient computation, scaling, performance optimizations, and real-time visualization leveraging NVIDIA GPUs.

We develop the prior density for the permeability and porosity fields ($K\mathrm{,\ }\varphi )$ using a Gaussian measure from a Gaussian process. We use this prior density to learn the \textit{PINO-\textit{CCR} } reservoir surrogate using Algorithm  \ref{alg1} with appropriate data and physics loss. Next this prior density is then used to learn the \textit{CCR} machine learning model required for the peaceman analytical well model using Algorithm  \ref{alg3} and then saved in the folder. For the sparse GP experts, we use the isotropic squared exponential covariance function with a variable number of inducing points $M_l$ based on the cluster sizes \cite{burt2020}. The pseudo-inputs $\tilde{x}_l$ are initialized via K-means and the GP hyperparameters are initialized based on the scale of the data. The number of experts $L$ is determined \textit{a priori} using the Bayesian information criterion (BIC) to compare the GMM clustering solutions of the rescaled $(y,x)$ across different values of $L$. The choice of GMM is motivated by allowing elliptically-shaped clusters in the rescaled space, while keeping cost low (e.g., if compared to the K-means). The GPytorch package \cite{gardner2021gpytorch} and Xgboost \cite{chen2016xgboost} were used to train the sparse Gaussian experts and Xgboost  classifier gating network respectively. We also learn the VCAE and DDIM models using Algorithm  \ref{alg4} and  \ref{alg5} and save the models in the folder.  

A selected number of realizations is then used for the inverse problem workload. In the inverse modeling we use Algorithm  \ref{alg8} where we  first forward using the prior developed \textit{PINO-\textit{CCR} }Reservoir simulator as shown in Algorithm  \ref{alg7} to get predicted well data of \textit{wopr, wwpr, wgpr} (well oil/water and gas production rates). The Peacemann analytical well model is described in Eqn.\ref{18a}
\begin{equation}\tag{18a}
\label{18a}
Q = \frac{2\pi K k_r h (P_{\text{avg}} - P_{\text{wf}})}{\mu B \ln \left(\frac{r_e}{r_w}\right) + s}
\end{equation}
\textbf{Where:}
\begin{itemize}
    \item $Q$ is the flow rate.
    \item $K k_r$ is the absolute permeability of the reservoir at the voxel location of the well.
    \item $k_r$ is the relative permeability which is a function of water and gas saturation.
    \item $h$ is the thickness of the reservoir.
    \item $P_{\text{avg}}$ is the average reservoir pressure.
    \item $P_{\text{wf}}$ is the bottom hole flowing pressure.
    \item $\mu$ is the viscosity of the fluid.
    \item $B$ is the formation volume factor.
    \item $r_e$ is the effective drainage radius.
    \item $r_w$ is the radius of the wellbore.
    \item $s$ is the skin factor.
\end{itemize}
This gives a novel implementation where we couple the \textit{CCR} algorithm and the \textit{PINO} approach to give a surrogate reservoir simulator called \textit{PINO-\textit{CCR} }, again Algorithm  \ref{alg7} explains this in detail.
\begin{equation}\tag{18b}
f_{\text{ccr}} (K, f_1(:,\theta_p), f_2(:,\theta_s), f_3(:,\theta_g); \theta_{\text{ccr}})
\end{equation}
We parametrise the permeability field by first using the encoder from Algorithm  \ref{alg4} to get the latent state $u_{x}$. We correct the Kalman Gain matrix with the compact support Gaspari cohn matrix. We then correct these quantities ($u_{x}\mathrm{,\ }FTM,\varphi )$ with \textit{a}REKI and we recover the spatial permeability ($K$) by applying the denoising function from Algorithm  \ref{alg5}. We repeat this cycle until a stopping criterion is met which is defined by the \textit{discrepancy principle} or the maximum number of iteration. The code for the numerical  implementation can be found at \url{https://github.com/NVIDIA/modulus-sym/tree/f59eba4d852a65cc80f703da754a87e51ba44d9d/examples/reservoir_simulation/Norne}
\section{Numerical Experiments}
\noindent The root-mean-square-error (RMSE) function for each ensemble member ($i$) is defined as
\begin{equation}\tag{19}
\mathrm{RMSE(}i\mathrm{)}={\left(\frac{1}{N}\sum^N_{k=1}{\sum^{N^k_{data}}_{j=1}{{\left(\frac{D^j_{obs}(k)-D^j_{sim,i}(k)}{{\sigma }_{n,j}}\right)}^2}}\right)}^{\frac{1}{2}}
\end{equation}
\noindent $N$:  Number of data assimilation time steps where measurements are assimilated (measurement times)
\noindent $N^k_{data}$: Number of data collected at each time step $k$
\noindent $i$: Ensemble member index
\noindent $k$: Time index
\noindent $j$: Metric or response (history matched metric or response)
\noindent $D^j_{obs}\left(k\right)$: Observed data metric for metrics \textit{j} (Data equivalent in state space ensemble)\textit{${}_{\ }$}at time step \textit{k}.
\noindent $D^j_{sim}\left(k\right)$: Simulated data from simulator for metrics \textit{j} (Data equivalent in state space ensemble)\textit{${}_{\ }$}at time step \textit{k}.
\noindent ${\sigma }_{n,j}$: Observed data standard deviation for metrics \textit{j} (Data equivalent in state space ensemble).
The result for the \textit{PINO} surrogate for the \textit{Norne} field is shown in Fig. (3 - 6). The  \textit{Norne} reservoir model is based on a real field black-oil model for an oil field in the Norwegian Sea \cite{Norne1,Norne2}. The grid is a faulted corner-point grid, with heterogeneous and anisotropic permeability. 100 training samples for the data loss requiring labeling and running the OPM-Flow simulator \cite{OPM} were used. The water flows from the injectors (downwards-facing arrows) towards the producers (upwards-facing arrows). The size of the reservoir computational voxel is $n_x, n_y, n_z = 46, 112, 22$. Three phases are considered (oil, gas, and water) and the wells (9 water injectors, 4 gas injectors, and 22 producers) are configured. The 22 producer wells have measurable quantities of oil rate, water rate, and gas rate and are controlled by bottom-hole-pressure. The 4 water injector wells have a measurable quantity of bottom hole pressure (BHP), controlled by injection rates. The reservoir is a sandstone Gaussian reservoir. 3,298 days of simulation are simulated. 100 realizations of the \textit{Norne} model that include variations in porosity, rock permeabilities, and fault transmissibility multipliers are used to generate the training and test data.

The left column of Fig. (3-6) is the responses from the \textit{PINO} surrogate, the middle column is the responses from the finite volume solver -\textit{Flow} and the right column is the difference between each response.
\begin{table}[h]
    \centering
    \caption{\textit{Norne} reservoir model simulation properties}
    \begin{tabular}{p{0.5\linewidth} p{0.4\linewidth}}
        \toprule
	\textbf{Property} & \textbf{Value} \\
        \midrule
        Grid Configuration & $46 \times 112 \times 22$ \\
        Grid size & $50 \times 50 \times 20$ ft \\
        \# of producers & 22 \\
        \# of injectors & 13 \\
        Simulation period & 3,298 days \\
        Integration step length & 100 days \\
        \# of integration steps & 30 \\
        Reservoir depth & 4,000 ft \\
        Initial Reservoir pressure & 1,000 psia \\
        Injector’s constraint & 500 STB/day \\
        Producer’s constraint & 100 psia \\
        Residual oil saturation & 0.2 \\
        Connate water saturation & 0.2 \\
        Petro-physical property distribution & Gaussian permeability/porosity \\
        Uncertain parameters to estimate & Distributed permeability, porosity, Fault transmissibility multiplier \\
        \bottomrule
    \end{tabular}
    \label{tab:norne}
\end{table}
\begin{algorithm}[H]
\caption{\textit{PINO} - \textit{CCR} Reservoir Simulator}
\small
\label{alg7}
\footnotesize 
\begin{algorithmic}[1]
\Require 
$X_0 = \{K, FTM, \varphi, P_{\text{ini}}, S_{\text{ini}}\} \in \mathbb{R}^{B_1 \times 1 \times D \times W \times H}$,
$f_1(:, \theta_p)$ - FNO pressure surrogate,
$f_2(:, \theta_s)$ - FNO Water saturation surrogate,
$f_3(:, \theta_g)$ - Gas saturation surrogate,
$f_{ccr}(:, \theta_{ccr})$ - \textit{CCR} Peaceman surrogate,
$N_{pr}$ - Number of producers,
$i_{pef}, j_{pef}, k_{pef}$ - locations/completions,
$N_x, N_y, N_z$ - size of the reservoir domain,
${shutin}_T$ - Boolean representing well shut-in state,
$T$ - Time index
\Ensure $Y_{out}$
\State Compute: $Y_{1p} = f_1(X_0; \theta_p)$, $Y_{1s} = f_2(X_0; \theta_s)$, $Y_{1g} = f_3(X_0; \theta_g)$
\State Construct: 
\begin{flalign*}
&X_1 = \left[ 
\begin{array}{c}
\text{if } {shutin}_T = \text{True} \begin{array}{c}
0, 0, 0, 0 \\
\end{array} \\
\text{if } {shutin}_T = \text{False} \begin{array}{c}
\frac{1}{N_z}\sum_{k=1}^{N_z} K_{i_{pef}, j_{pef}, k}, \\
\frac{1}{N_{pr}}\sum_{n=1}^{N_{pr}}\frac{1}{N_z}\sum_{k=1}^{N_z} f_1(X_1, \theta_p)_{i_{pef}n, j_{pef}n, k}, \\
\frac{1}{N_z}\sum_{k=1}^{N_z} f_2(X_1, \theta_s)_{i_{pef}n, j_{pef}n, k}, \\
\frac{1}{N_z}\sum_{k=1}^{N_z} f_3(X_1, \theta_g)_{i_{pef}n, j_{pef}n, k}, \\
1 - \left(\frac{1}{N_z}\sum_{k=1}^{N_z} f_2(X_1, \theta_s)_{i_{pef}n, j_{pef}n, k} + \frac{1}{N_z}\sum_{k=1}^{N_z} f_3(X_1, \theta_g)_{i_{pef}n, j_{pef}n, k}\right)
\end{array}
\end{array}
\right] &&
\end{flalign*}
\State Compute: $Y_{out} = f_{ccr}(X_1; \theta_{ccr}) \in \mathbb{R}^{B_1 \times T \times 3N_{pr}}$  \Comment{\textit{wopr, wwpr, wgpr}}
 \State \textbf{Output:} $Y_{out}$
\end{algorithmic}
\end{algorithm}
\vspace{-0.5cm}
\begin{algorithm}[H]
\caption{Adaptive Regularized Kalman Inversion (aREKI) using PINO-CCR and VCAE/DDIM exotic priors}
\small
\label{alg8}
\begin{algorithmic}[1]
\Require
\begin{itemize}
    \item$X_0 = \{K, FTM, \varphi, P_{\text{ini}}, S_{\text{ini}}\} \in \mathbb{R}^{B_1 \times 1 \times D \times W \times H}$\Comment{Initial ensemble}
    \item$f_1(:, \theta_p)$\Comment{\textit{PINO} pressure surrogate}
    \item$f_2(:, \theta_s)$ \Comment{\textit{PINO} Water saturation surrogate}
    \item$f_3(:, \theta_g)$ \Comment{\textit{PINO} Gas saturation surrogate}
    \item$f_{ccr}(:, \theta_{ccr})$ \Comment{\textit{CCR} Peaceman surrogate}
    \item$N_{pr}$ \Comment{Number of producers}
    \item$i_{pef}, j_{pef}, k_{pef}$ \Comment{locations/completions}
    \item$N_x, N_y, N_z$\Comment{ size of the reservoir domain}
    \item${shutin}_T$ \Comment{Boolean representing well shut-in state}
    \item$T$ \Comment{Time index}
    \item $y$:\Comment{Measurement data}
    \item $\mathrm{\Gamma}$\Comment{Covariance matrix of measurement errors}
    \item $iter$\Comment{Maximum number of iterations}
\end{itemize}
\State Initialize: $s_n = 0$, $n = 0$
\While{$(s_n < 1) \text{ and } (n \leq iter)$}
   \State  Construct $X^{(j)}_n = \{K^{(j)}_{n}, FTM^{(j)}_{n} \varphi^{(j)}_{n}, P_{\text{ini}}, S_{\text{ini}}\}$ \Comment{tensor ensemble}
\State Compute predictions: $G(X^{(j)}_n) = f_{ccr}\left( X^{(j)}_n, f_1(:, \theta_p), f_2(:, \theta_s), f_3(:, \theta_g);\theta_{ccr} \right)$ for $j = 1, \dots, J$ \Comment{Refer to Algorithm \ref{alg7}}
    \State Encode $K_{n}$ to get latent variables: $u_{xn}, \mu_{n}, \sigma_{n} = \text{Encoder}_\phi(K_{n})$\Comment{Refer to Algorithm \ref{encoderr}}
    \State Compute residuals: $\mathrm{\Phi}_n \equiv \left\{\frac{1}{2}\left\|\mathrm{\Gamma}^{-\frac{1}{2}}(y - G(X^{(j)}_n))\right\|^2_2\right\}^J_{j=1}$
    \State Compute mean and variance of residuals: $\overline{\mathrm{\Phi}_n}$, $\sigma^2_{\mathrm{\Phi}_n}$
    \State Adjust step size: $\alpha_n = \frac{1}{\min \left\{\max \left\{\frac{L_{(y)}}{2\overline{\mathrm{\Phi}_n}}, \sqrt{\frac{L_{(y)}}{2\sigma^2_{\mathrm{\Phi}_n}}}\right\}, 1 - s_n \right\}}$
    \State Update accumulated step: $s_n = s_n + \frac{1}{\alpha_n}$
\State Construct the parameters to recover \( u^{(j)}_n = \left[ \begin{array}{c}
u_{xn} \\
\varphi_n \\
\text{FTM}_n
\end{array} \right]^j \).

    \State Compute cross-covariance matrices:
    \State \hspace{\algorithmicindent} $C^{GG}_n= \frac{1}{J-1}\sum^J_{j=1}\left(G(X^{(j)}_n) - \overline{G(X^{(j)}_n)}\right) \otimes \left(G(X^{(j)}_n) - \overline{G(X^{(j)}_n)}\right)$
    \State \hspace{\algorithmicindent} $C^{uG}_n= \frac{1}{J-1}\sum^J_{j=1}\left(u^{(j)}_n - \overline{u^{(j)}_n}\right) \otimes \left(G(X^{(j)}_n) - \overline{G(X^{(j)}_n)}\right)$
    \State Update ensemble: 
    \State \hspace{\algorithmicindent} $u^{(j)}_{n+1} = u^{(j)}_n + C^{uG}_n\left(C^{GG}_n + \alpha_n \mathrm{\Gamma}\right)^{-1}(y + \sqrt{\alpha_n}\xi_n - G(X^{(j)}_n))$
  \For{$t = 1$ to $T$}
    \State $\hat{u}^{(j)}_{x,t-1} = f_\theta(u^{(j)}_{x,t}, t)$ \Comment{Refer to Algorithm \ref{alg5}, where $f_\theta$ is the denoising DDIM function}
    \State $u^{(j)}_{x,t} = \hat{u}^{(j)}_{x,t-1}$
    \If{$t == T$}
      \State $K^{(j)}_{n} = u^{(j)}_{x,T}$
    \EndIf
  \EndFor
    \State Increment counter: $n \leftarrow n + 1$
\EndWhile
\State \textbf{Output:} $\{K^{(j)}_n, \varphi^{(j)}_n, \text{FTM}^{(j)}_n\}^J_{j=1}$: \Comment{Updated permeability,porosity and fault multiplier fields}
\end{algorithmic}
\end{algorithm}
\begin{figure}[htbp!]
\centering
\includegraphics[width=6.27in, height=6.27in, keepaspectratio]{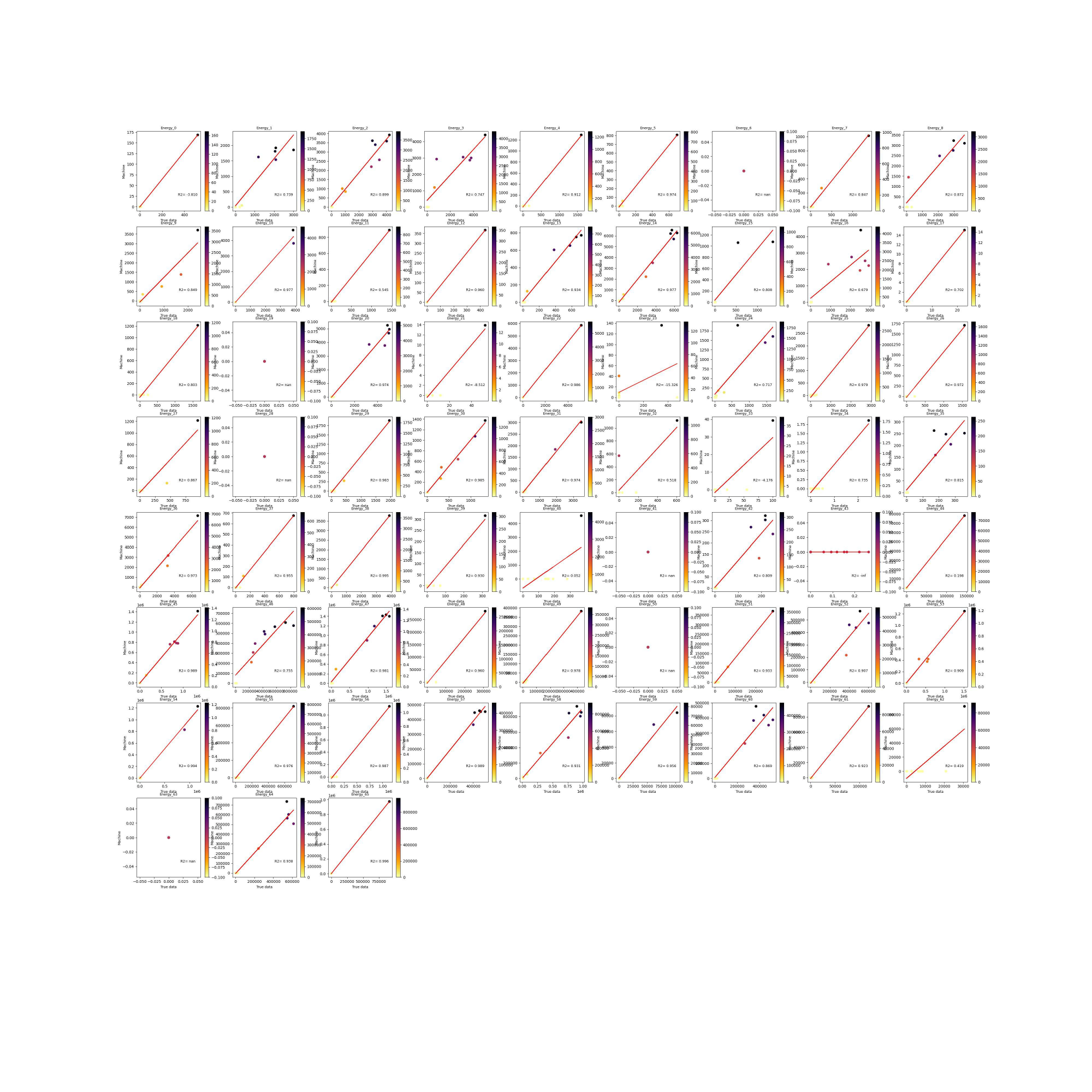}
\caption{\textit{Forwarding of the Norne Field.} \(N_x=46, N_y=112, N_z=22\). \textit{Peaceman well model surrogate with \textit{\textit{CCR}}. The 66 tile indicates each of the 66 outputs for the wopr, wwpr, wgpr for the 22 wells. The accuracies are computed on the unseen validation text set.}}
\end{figure}
\noindent \textit{}
\noindent Figure 3 - 6 shows the dynamic responses for the pressure, water saturation and gas saturation fields for the PINO surrogate simulator and the numerical solver. Figure 7(a-c) shows the well response match between the \textit{PINO-\textit{CCR}} surrogate and the actual true data. Figure 9 - 12 shows the results for the inverse modeling. Figure 9 - 12  is the ultimate target in such reservoir characterization task, and it shows the well responses for the each of the percentiles (P10, P50,P90) match to the true model.  
\begin{figure}[htbp!]
\centering
\includegraphics[width=6.25in, height=6.25in, keepaspectratio]{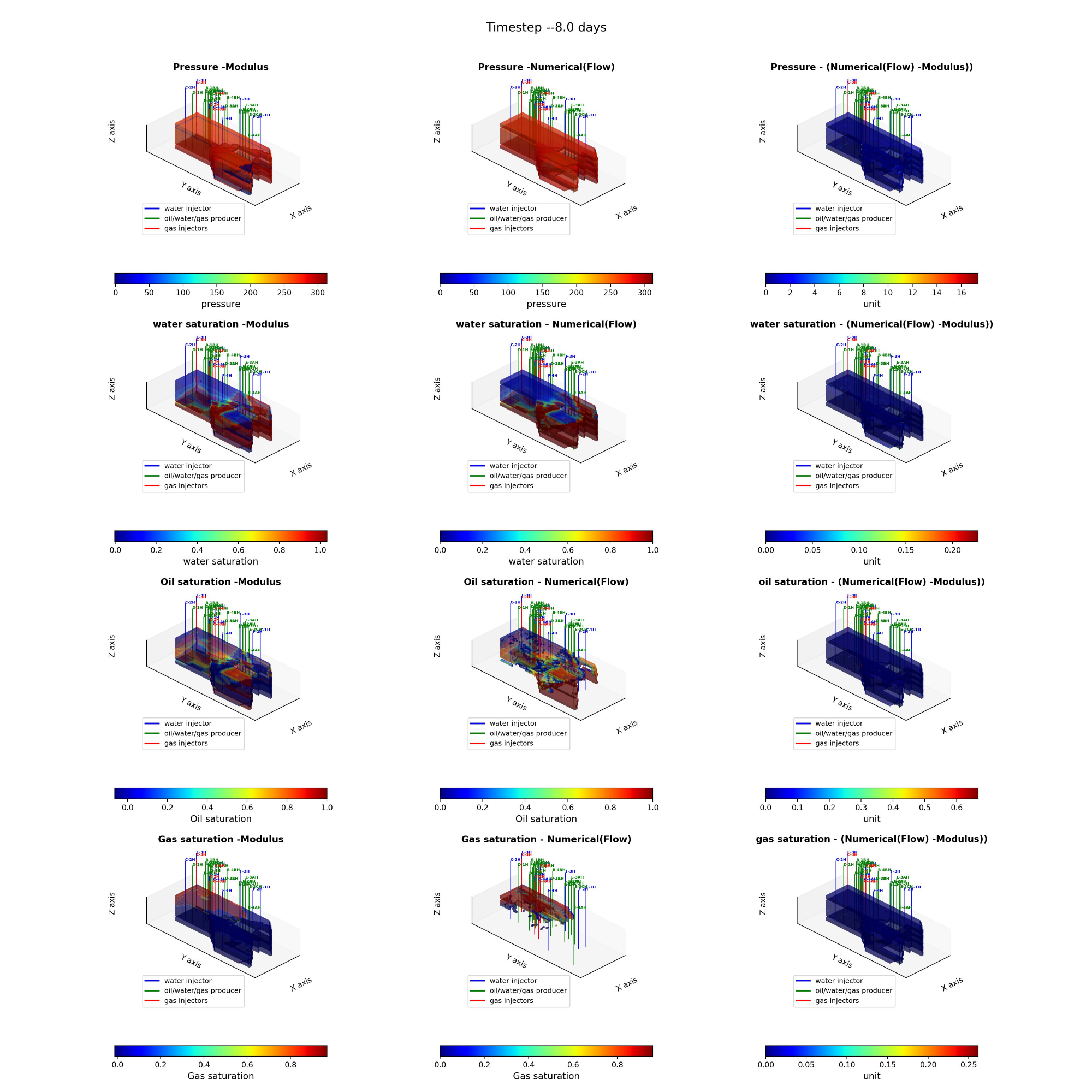} 
\caption{\textit{Forwarding of the Norne Field.} \(N_x=46, N_y=112, N_z=22\). \textit{At Time = 8 days. Dynamic properties comparison between the pressure, water saturation, oil saturation and gas saturation field between Nvidia Modulus's PINO surrogate (left column), Numerical solver reservoir simulator (middle column) and the difference between both approaches (last column). They are 22 oil/water/gas producers (green), 9 water injectors (blue) and 4 gas injectors) red. We can see good concordance between the surrogate's prediction and the numerical reservoir simulator (Numerical solver).}}
\end{figure}
\begin{figure}[htbp!]
\centering
\includegraphics[width=6.25in, height=6.25in, keepaspectratio]{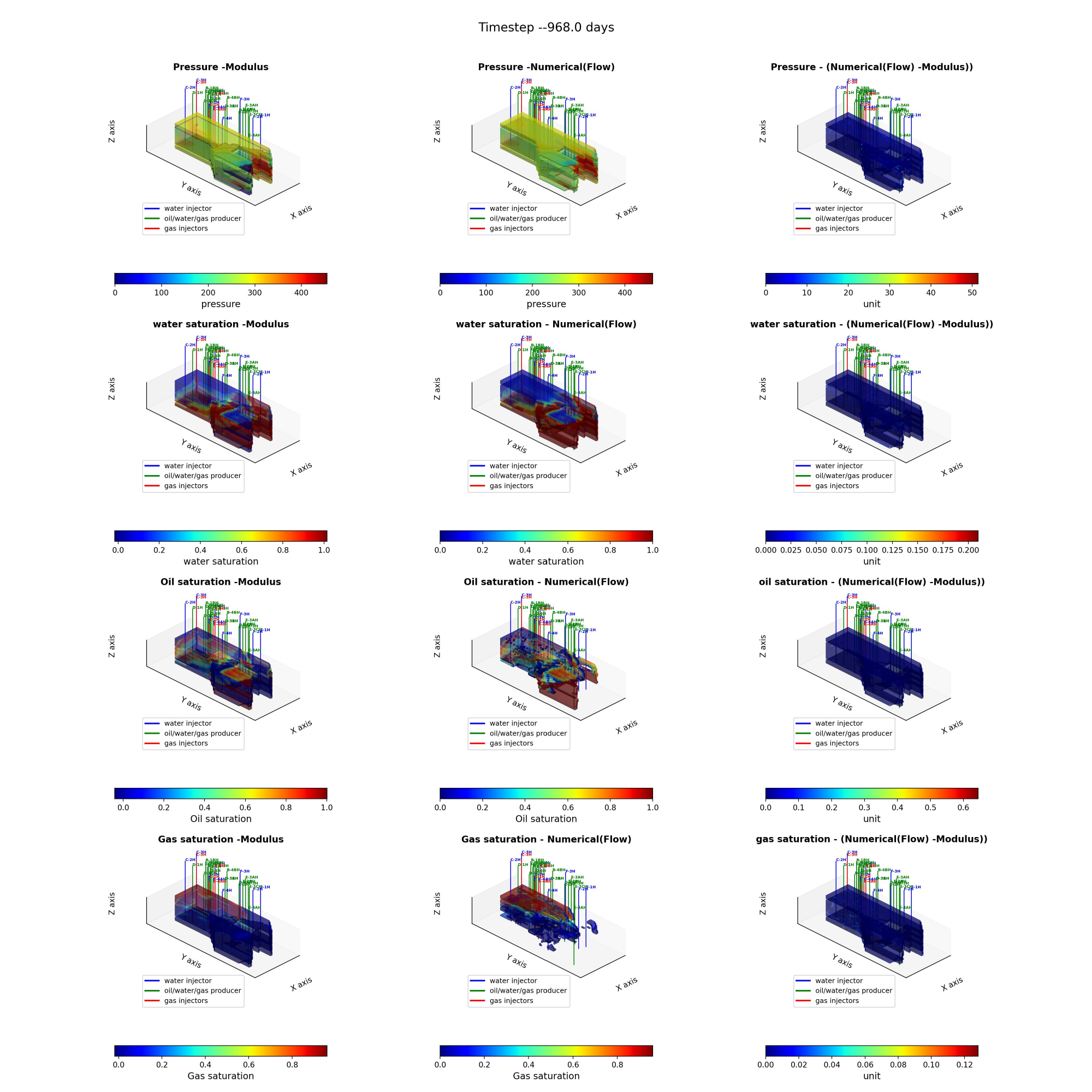} 
\caption{\textit{Forwarding of the Norne Field.} \(N_x=46, N_y=112, N_z=22\). \textit{At Time = 968 days. Dynamic properties comparison between the pressure, water saturation, oil saturation and gas saturation field between Nvidia Modulus's PINO surrogate (left column), Numerical solver reservoir simulator (middle column) and the difference between both approaches (last column). They are 22 oil/water/gas producers (green), 9 water injectors (blue) and 4 gas injectors) red. We can see good concordance between the surrogate's prediction and the numerical reservoir simulator (Numerical solver).}}
\end{figure}
\begin{figure}[htbp!]
\centering
\includegraphics[width=6.25in, height=6.25in, keepaspectratio]{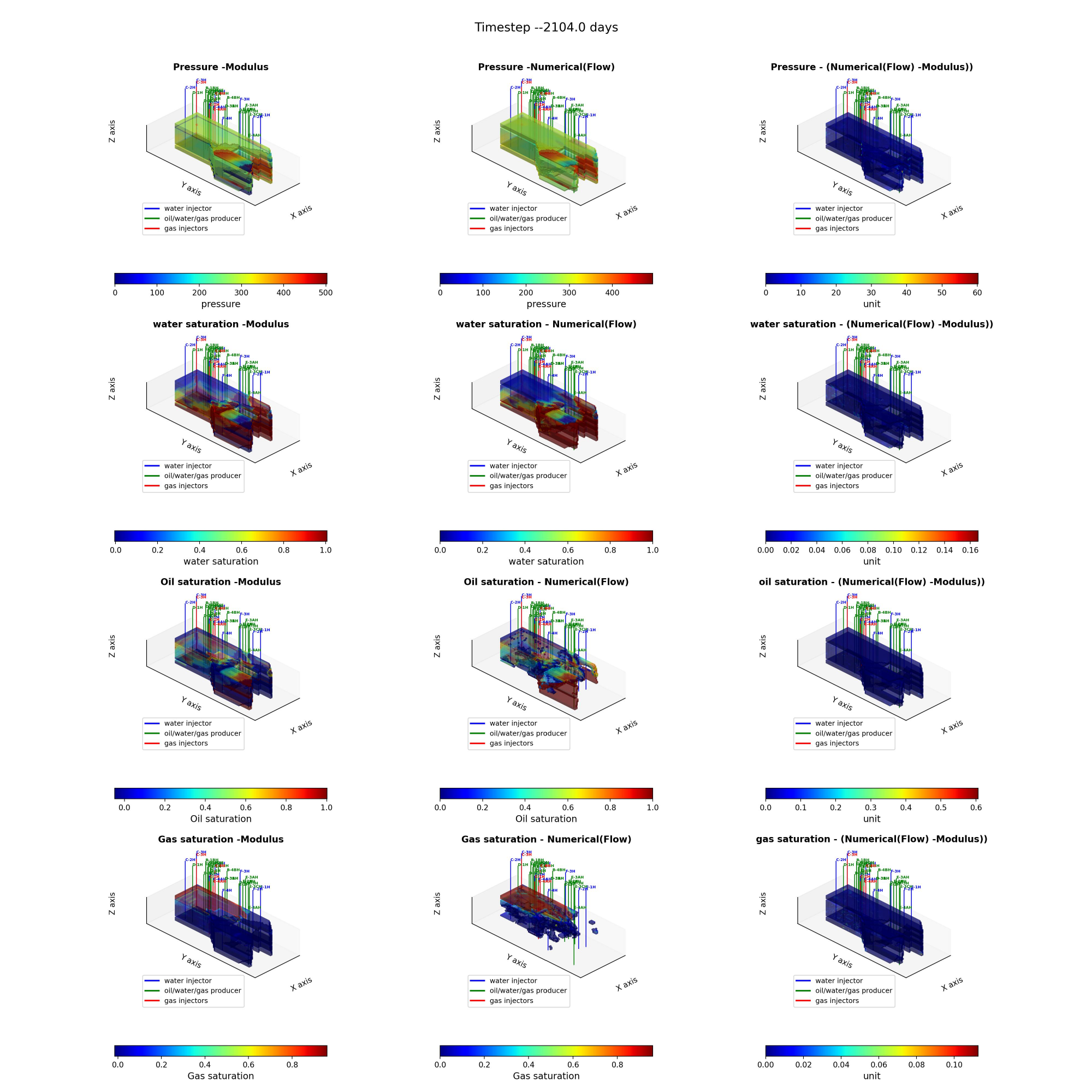} 
\caption{\textit{Forwarding of the Norne Field.} \(N_x=46, N_y=112, N_z=22\). \textit{At Time = 2104 days. Dynamic properties comparison between the pressure, water saturation, oil saturation and gas saturation field between Nvidia Modulus's PINO surrogate (left column), Numerical solver reservoir simulator (middle column) and the difference between both approaches (last column). They are 22 oil/water/gas producers (green), 9 water injectors (blue) and 4 gas injectors) red. We can see good concordance between the surrogate's prediction and the numerical reservoir simulator (Numerical solver).}}
\end{figure}
\begin{figure}[htbp!]
\centering
\includegraphics[width=6.25in, height=6.25in, keepaspectratio]{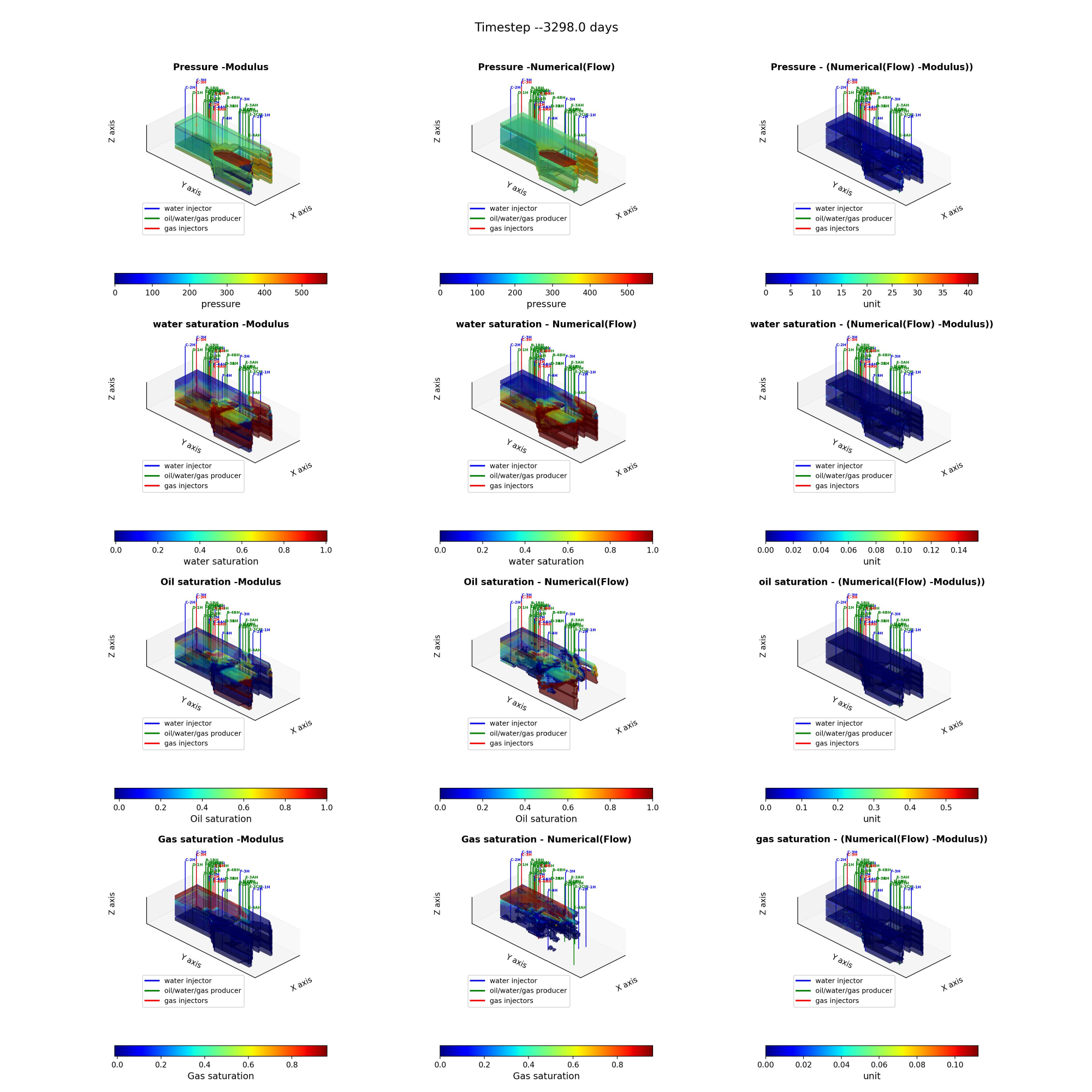} 
\caption{\textit{Forwarding of the Norne Field.} \(N_x=46, N_y=112, N_z=22\). \textit{At Time = 3298 days. Dynamic properties comparison between the pressure, water saturation, oil saturation and gas saturation field between Nvidia Modulus's PINO surrogate (left column), Numerical solver reservoir simulator (middle column) and the difference between both approaches (last column). They are 22 oil/water/gas producers (green), 9 water injectors (blue) and 4 gas injectors) red. We can see good concordance between the surrogate's prediction and the numerical reservoir simulator (Numerical solver).}}
\end{figure}
\begin{figure}[htbp!]
    \centering
    \begin{tabular}{cc}
        \begin{subfigure}[t]{0.45\textwidth}
            \centering
            \includegraphics[width=\linewidth, height=\linewidth, keepaspectratio]{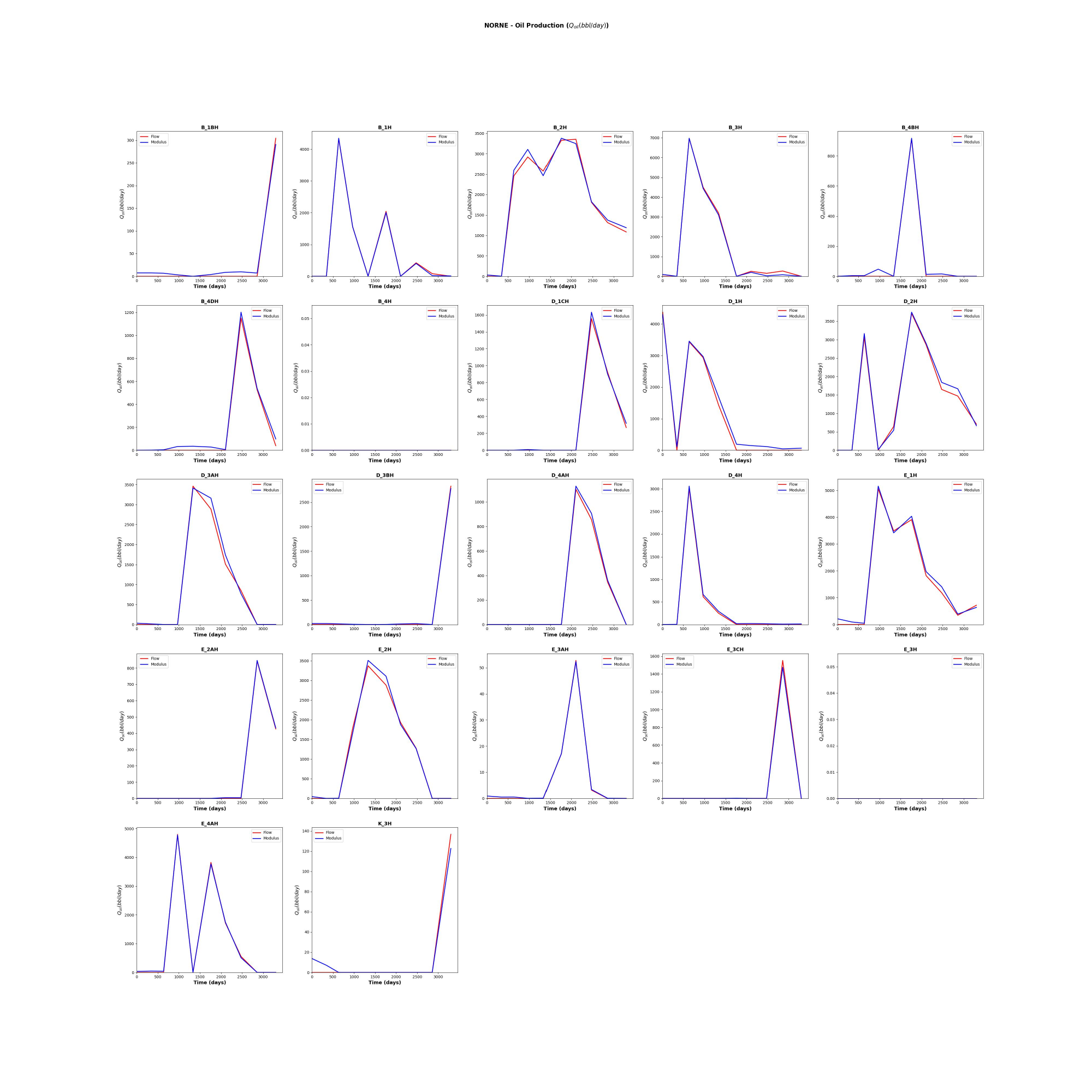}
            \caption{\textit{Forwarding of the Norne Field Comparison between using the. (red) True model, (blue) PINO-\textit{CCR} model. Outputs from peaceman machine for oil production rate,} $Q_o \frac{(stb)}{(day)}$ \textit{ from the 22 producers. They are 22 oil/water/gas producers , 9 water injectors and 4 gas injectors.}}
        \end{subfigure} & 
        \begin{subfigure}[t]{0.45\textwidth}
            \centering
            \includegraphics[width=\linewidth, height=\linewidth, keepaspectratio]{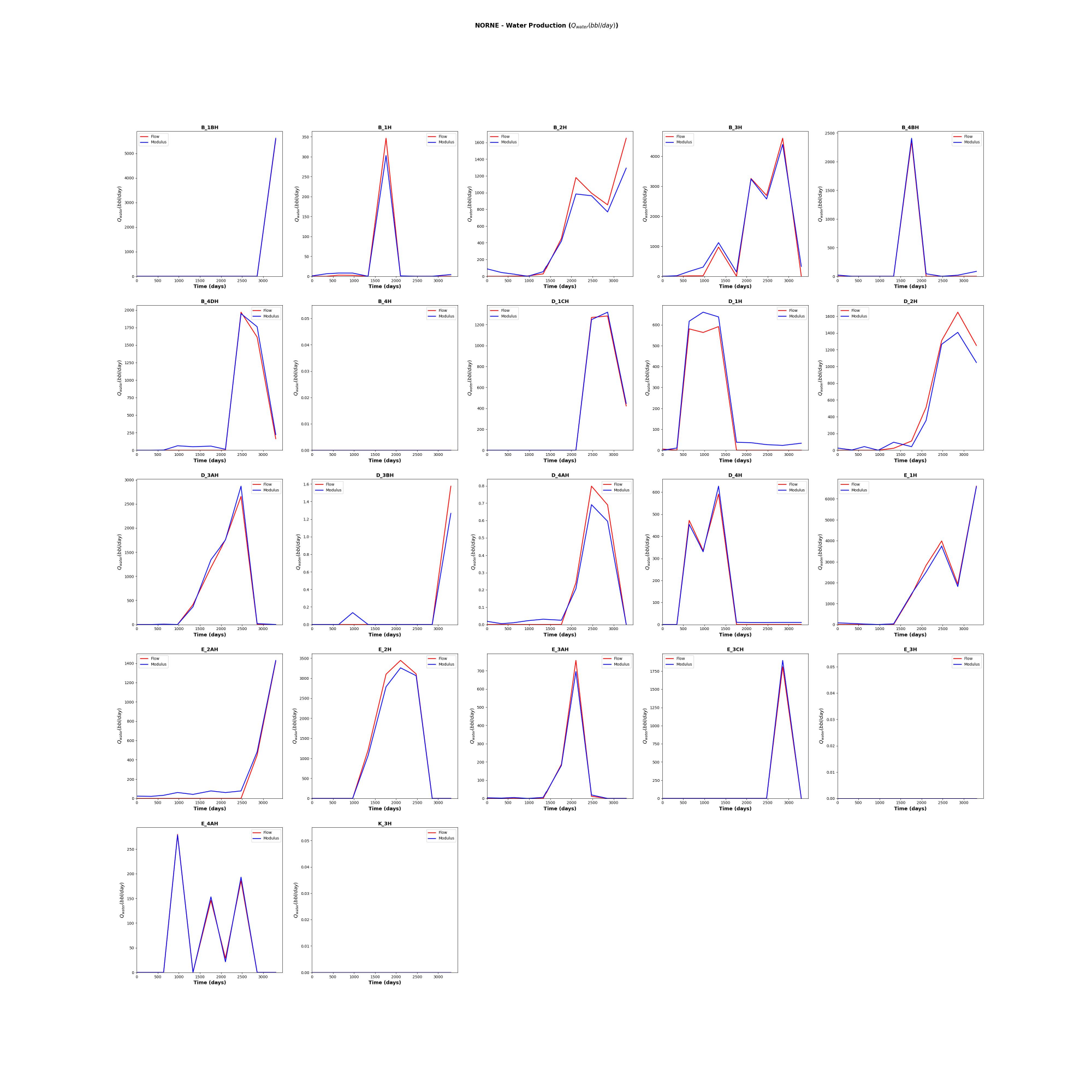}
            \caption{\textit{Forwarding of the Norne Field Comparison between using the. (red) True model, (blue) PINO-\textit{CCR} model. Outputs from peaceman machine for water production rate,} $Q_w \frac{(stb)}{(day)}$ \textit{ from the 22 producers.They are 22 oil/water/gas producers , 9 water injectors and 4 gas injectors.}}
        \end{subfigure} \\ 
        \begin{subfigure}[t]{0.45\textwidth}
            \centering
            \includegraphics[width=\linewidth, height=\linewidth, keepaspectratio]{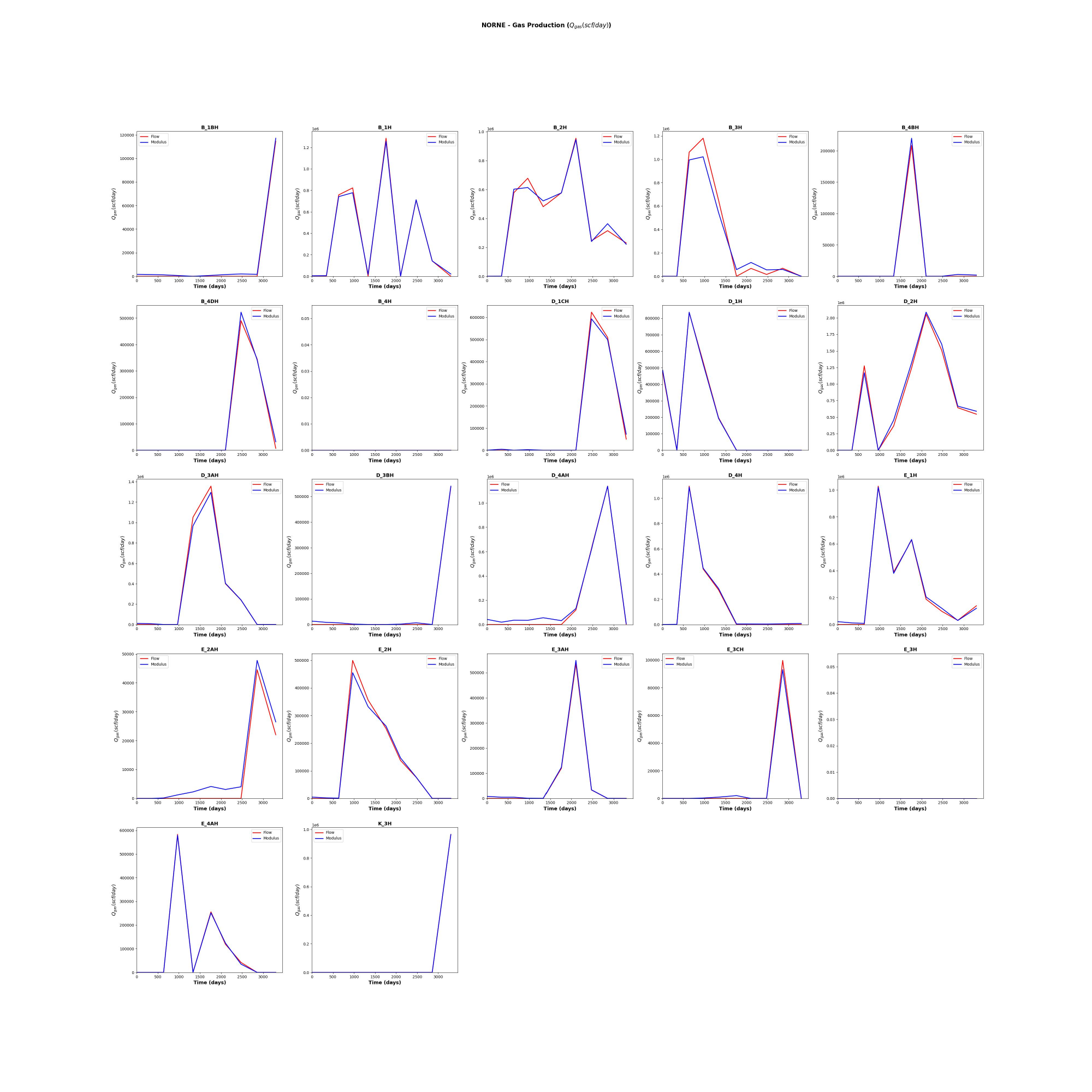}
            \caption{\textit{Forwarding of the Norne Field Comparison between using the. (red) True model, (blue) PINO-\textit{CCR} model. Outputs from peaceman machine for gas production rate,} $Q_g \frac{(scf)}{(day)}$ \textit{ from the 22 producers. They are 22 oil/water/gas producers , 9 water injectors and 4 gas injectors.}}
        \end{subfigure} & 
        \begin{subfigure}[t]{0.45\textwidth}
            \centering
            \vspace{0pt} 
        \end{subfigure}
    \end{tabular}
    \caption{\textit{Overall forwarding of the Norne Field using True model (red) and PINO-\textit{CCR} model (blue). Outputs from Peaceman machine for oil, water, and gas production rates from the 22 oil/water/gas producers , 9 water injectors and 4 gas injectors.}}
\end{figure}
\begin{figure}[htbp!]
\centering
\includegraphics[width=4.25in, keepaspectratio]{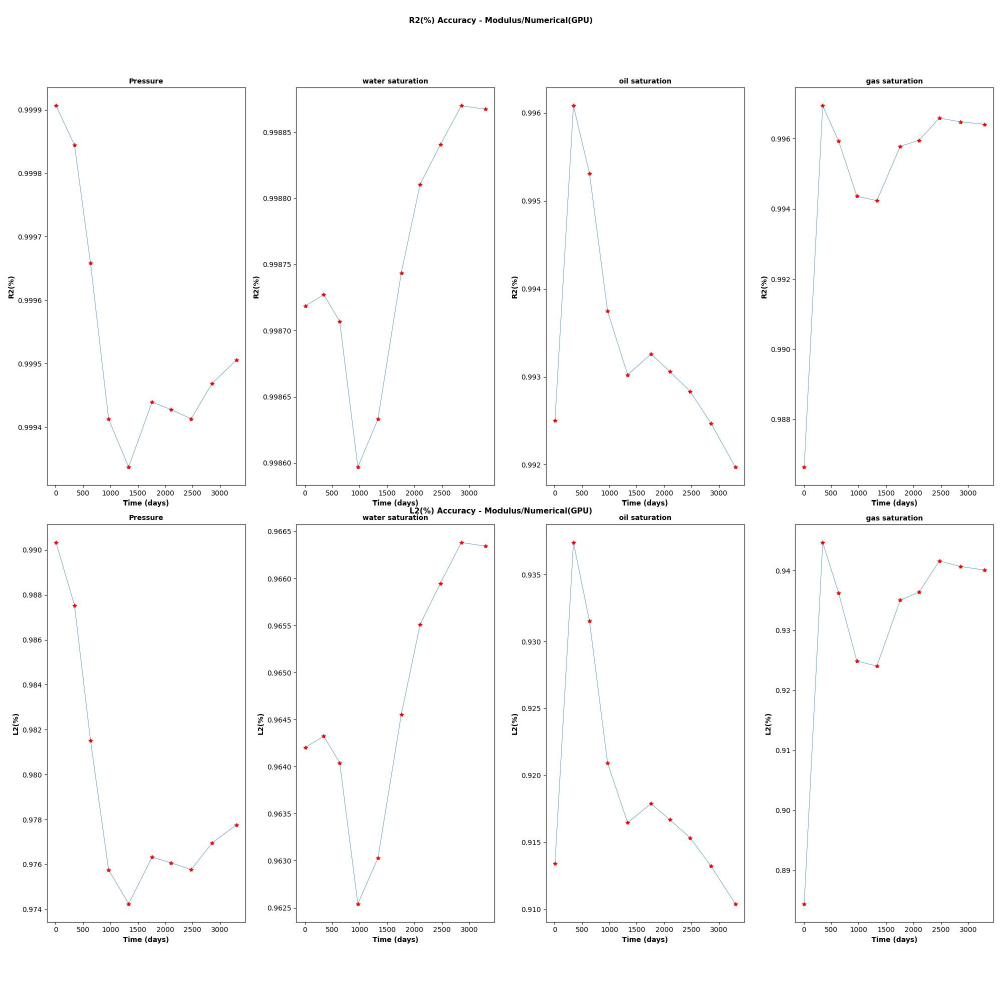} 
\caption{\textit{Forwarding of the Norne Field (First row): }$R_2\ $\textit{accuracy, (bottom-row) }$L_2$\textit{ accuracy. Column 1-4 denotes the pressure, water saturation, oil saturation and gas saturation field respectively.}}
\end{figure}
\begin{figure}[H]
    \centering
    \begin{subfigure}[b]{0.3\textwidth}
        \centering
        \includegraphics[width=\textwidth, keepaspectratio]{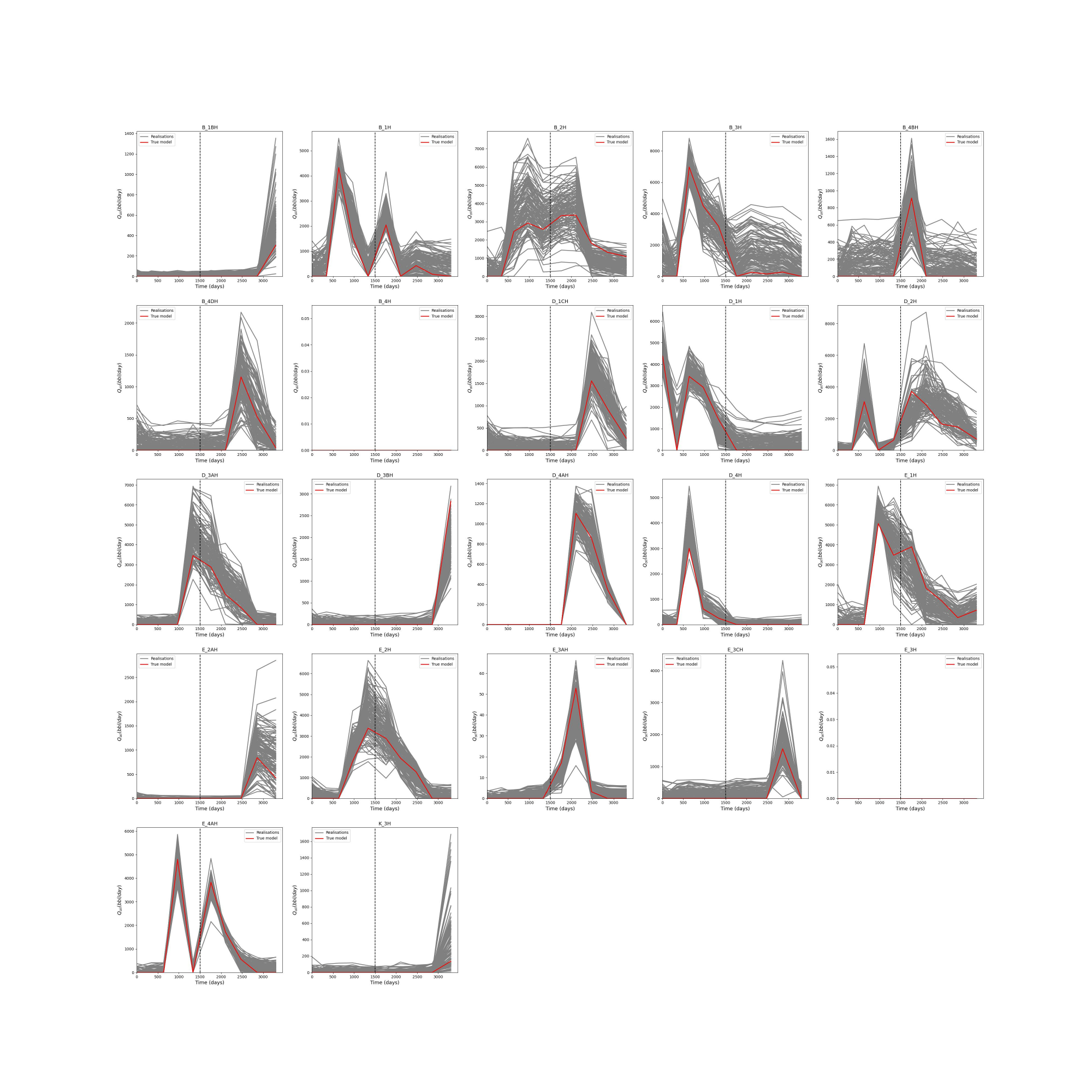}
        \caption{\textit{prior ensemble for oil production rate, $Q_o$ (stb/day).}}
        \label{fig:image1}
    \end{subfigure}
    \hfill
    \begin{subfigure}[b]{0.3\textwidth}
        \centering
        \includegraphics[width=\textwidth, keepaspectratio]{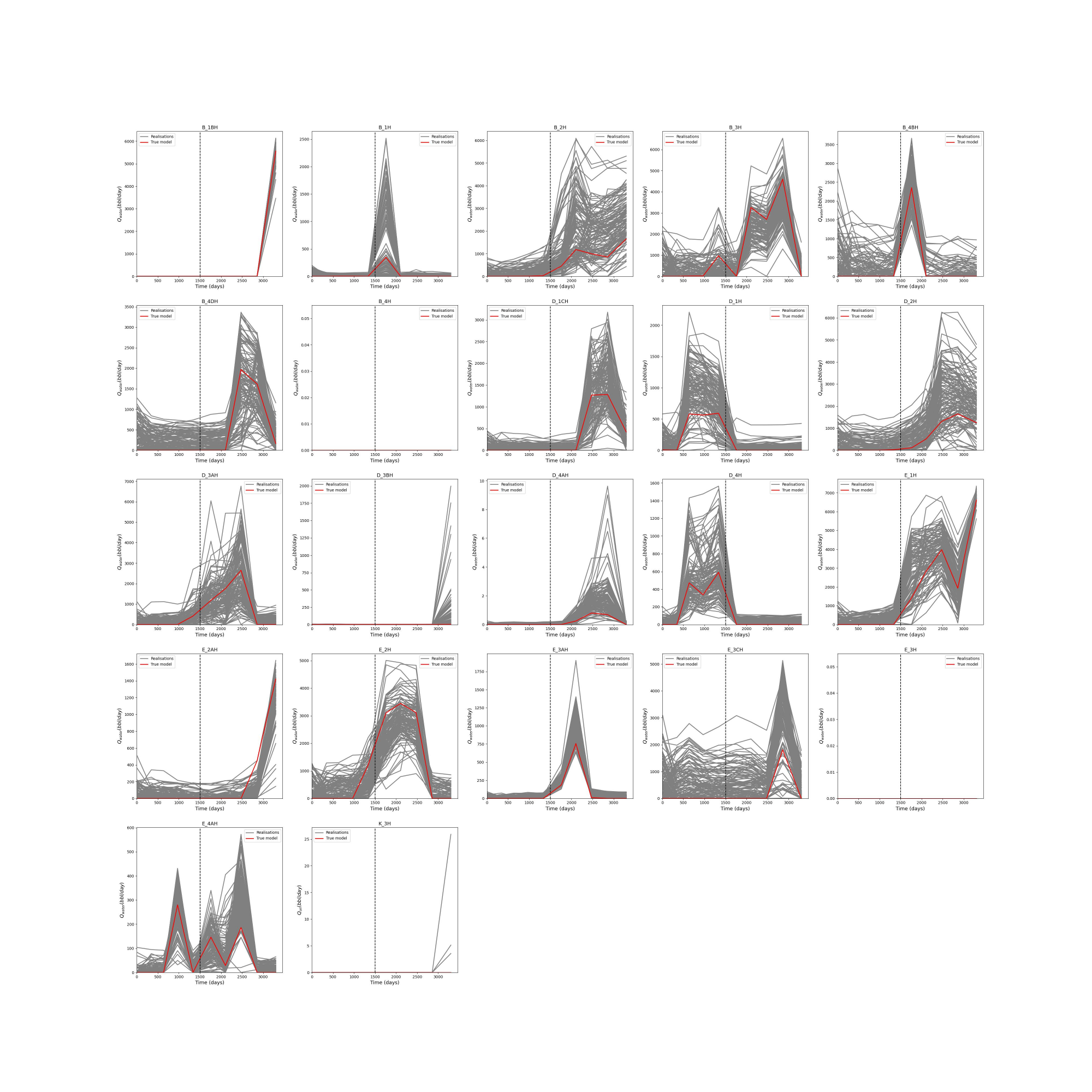}
        \caption{\textit{ prior ensemble for water production rate, $Q_w$ (stb/day).}}
        \label{fig:image2}
    \end{subfigure}
    \hfill
    \begin{subfigure}[b]{0.3\textwidth}
        \centering
        \includegraphics[width=\textwidth, keepaspectratio]{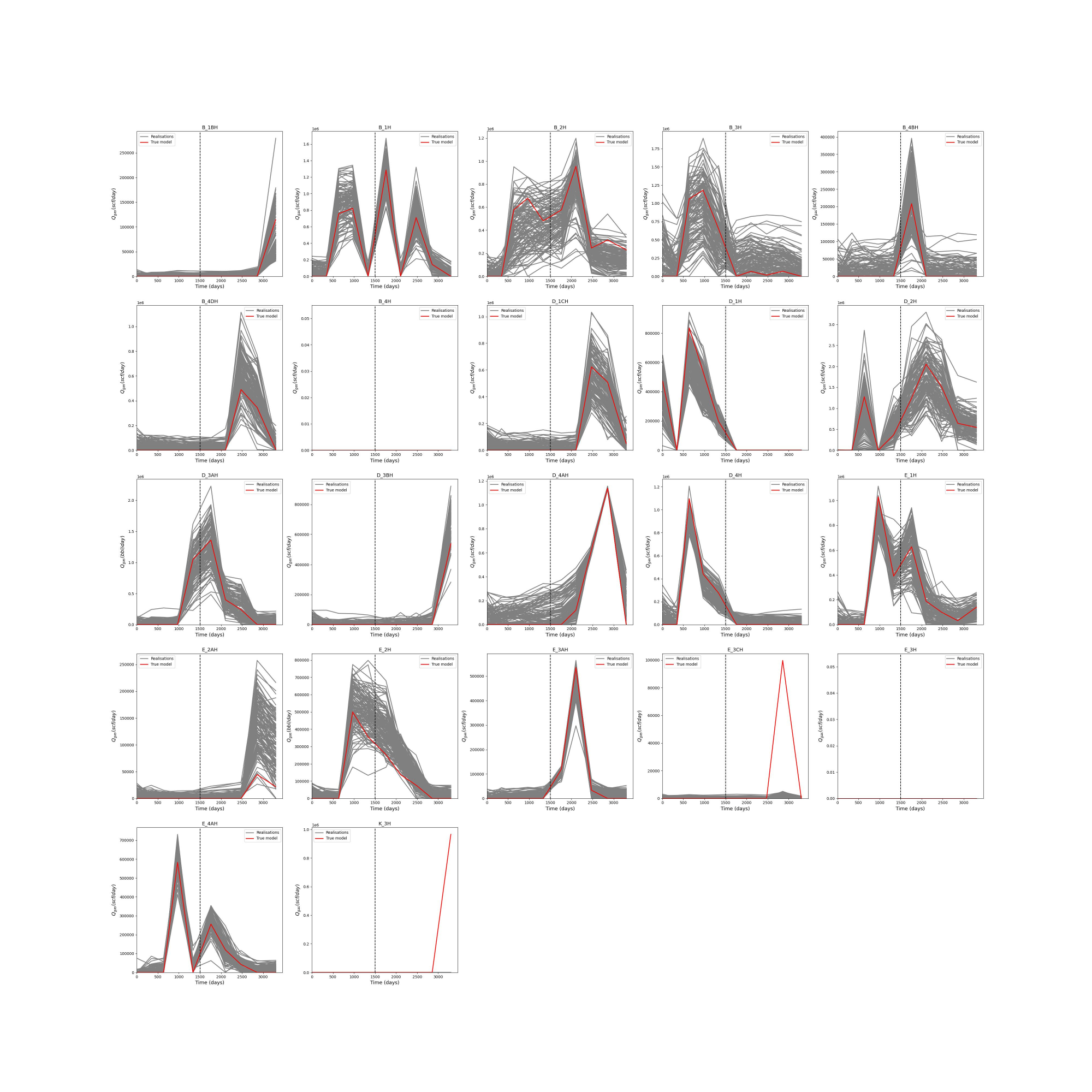}
        \caption{\textit{ prior ensemble for gas production rate, $Q_g$ (scf/day).}}
        \label{fig:image3}
    \end{subfigure}
    \vspace{0.5cm}   
    \begin{subfigure}[b]{0.3\textwidth}
        \centering
        \includegraphics[width=\textwidth, keepaspectratio]{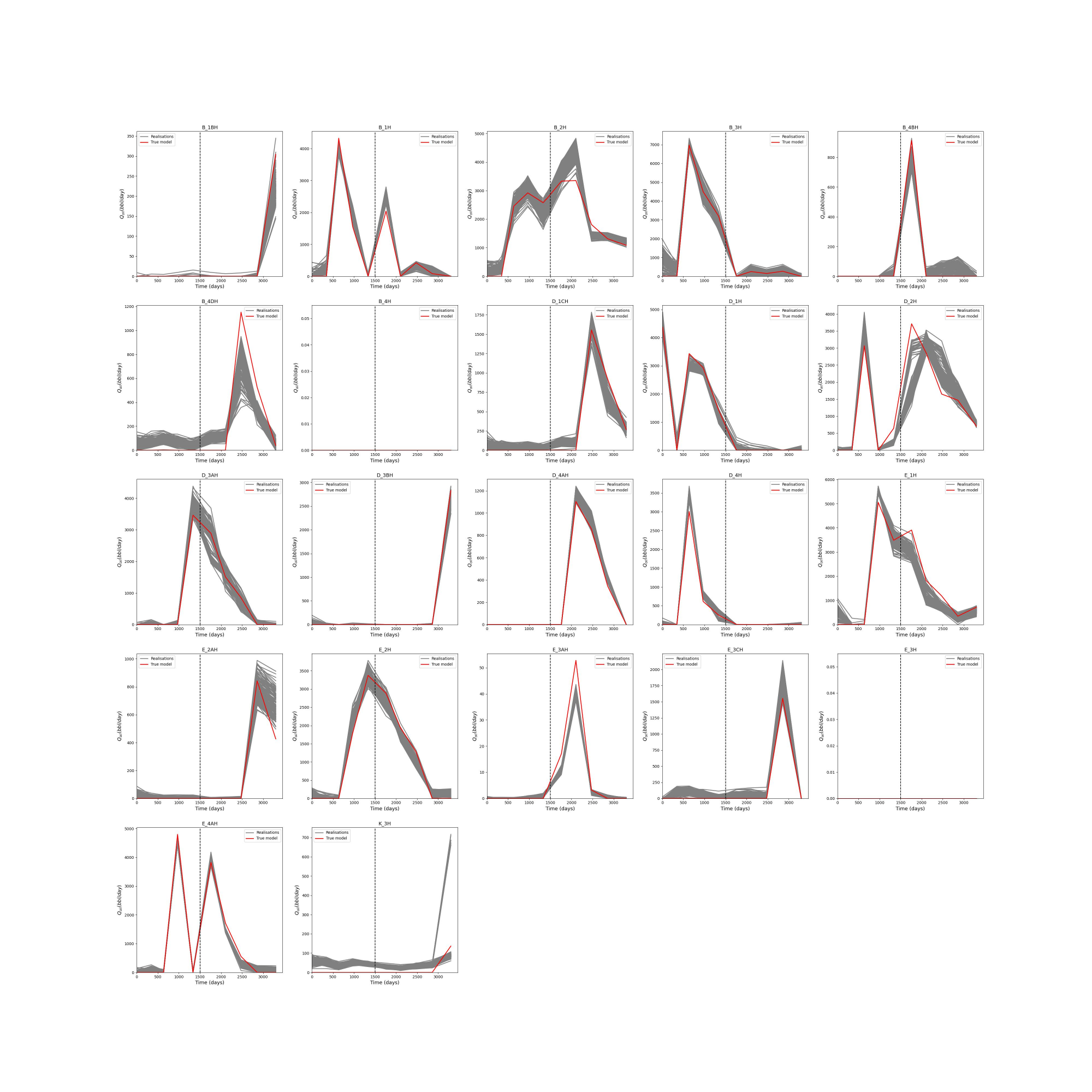}
        \caption{\textit{posterior ensemble for oil production rate, $Q_o$ (stb/day).}}
        \label{fig:image4}
    \end{subfigure}
    \hfill
    \begin{subfigure}[b]{0.3\textwidth}
        \centering
        \includegraphics[width=\textwidth, keepaspectratio]{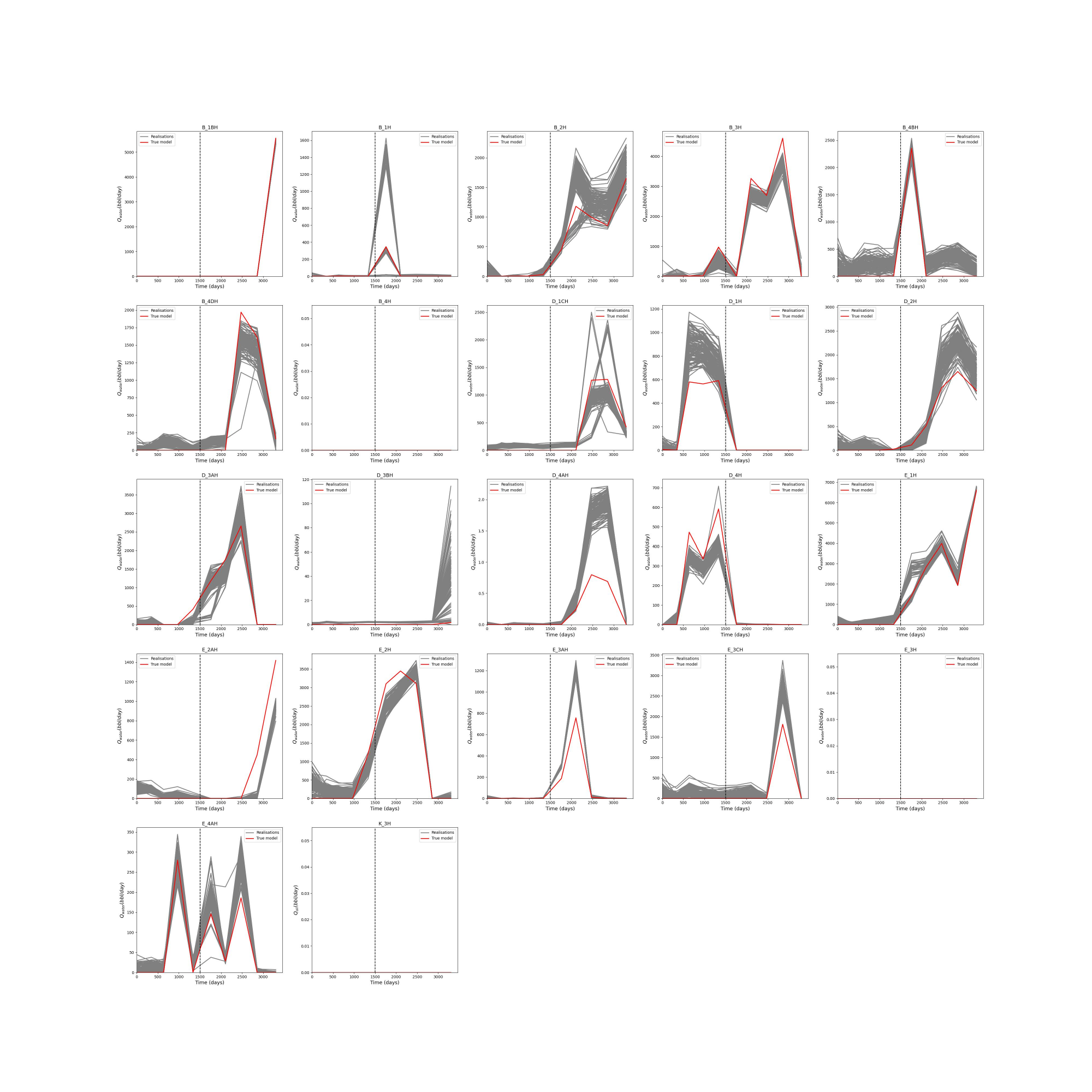}
        \caption{\textit{posterior ensemble for water production rate, $Q_w$ (stb/day).}}
        \label{fig:image5}
    \end{subfigure}
    \hfill
    \begin{subfigure}[b]{0.3\textwidth}
        \centering
        \includegraphics[width=\textwidth, keepaspectratio]{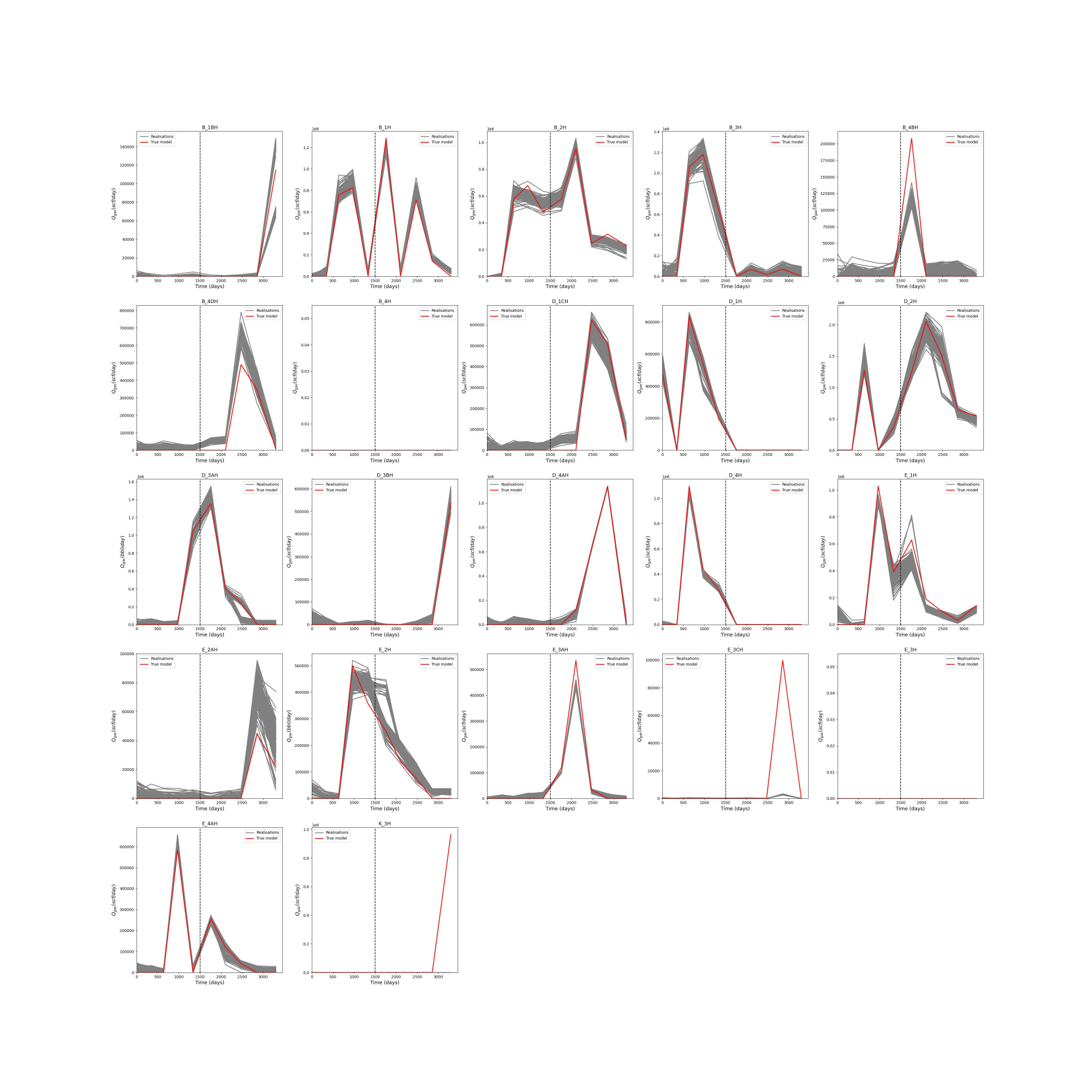}
        \caption{\textit{ posterior ensemble for gas production rate, $Q_g$ (scf/day).}}
        \label{fig:image6}
    \end{subfigure}

    \caption{\textit{Inverse modeling with covariance localization and \textit{a}REKI}}
    \label{fig:all_images}
\end{figure}
\begin{figure}[htbp!]
\centering
\includegraphics[width=4.25in, height=6.25in, keepaspectratio]{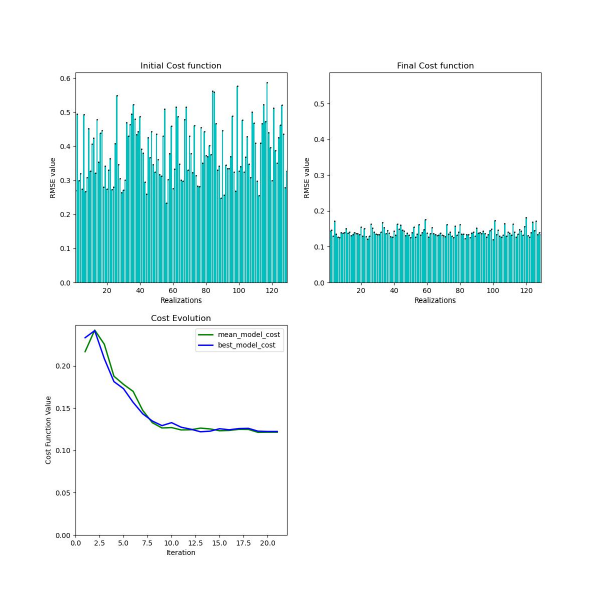} 
\caption{\textit{Cost function evolution. (Top -left) prior ensemble RMSE cost, (Top-right) posterior ensemble RMSE cost, (Bottom-left) RMSE cost evolution between the MAP model (blue) and the MLE model (green).}}
\end{figure}
\begin{figure}[htbp!]
\centering
\includegraphics[width=4.25in, height=4.25in, keepaspectratio]{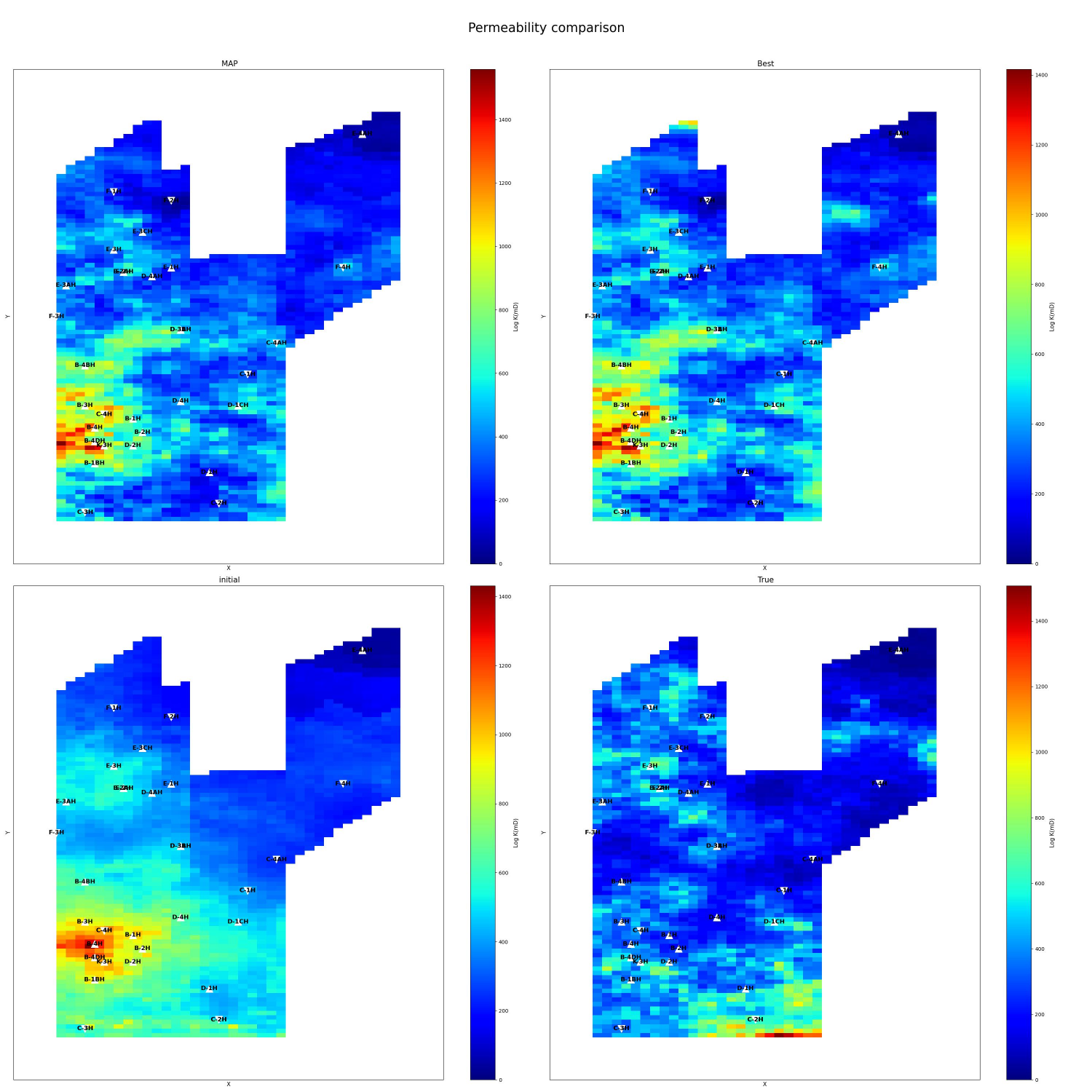} 
\caption{\textit{Comparison of permeability field reconstruction (1st column) prior, (2nd column) MLE estimate (posterior) (third column) MAP estimate (posterior) and (4th-column) True Model. The method used is the -- \textit{a}REKI + Covariance localization.}}
\end{figure}
\begin{figure}[htbp!]
    \centering
    \begin{subfigure}[b]{0.45\textwidth}
        \centering
        \includegraphics[width=\textwidth, keepaspectratio]{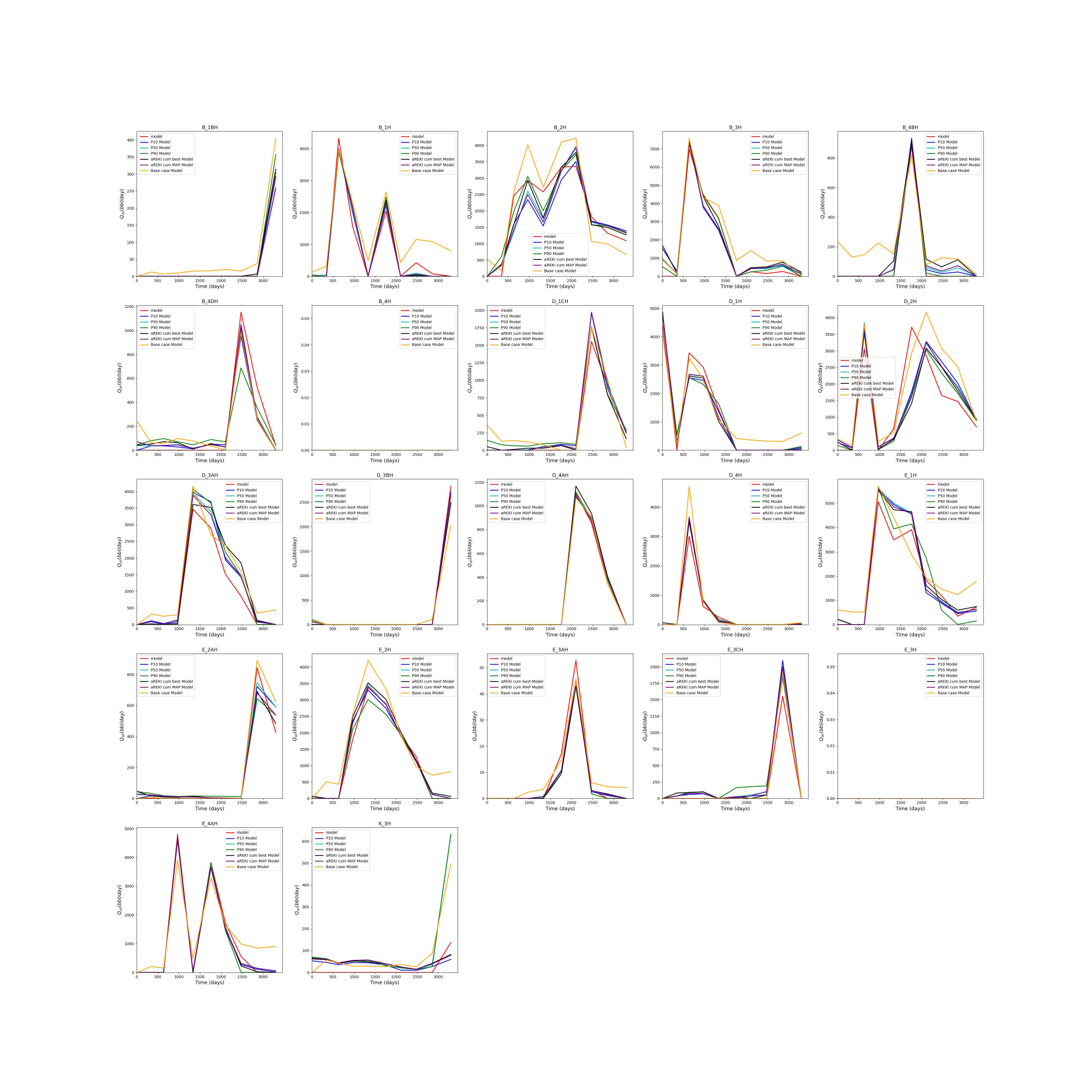}
        \caption{\textit{$Q_o\frac{(stb)}{(day)}$ }}
        \label{fig:oil}
    \end{subfigure}
    \hfill
    \begin{subfigure}[b]{0.45\textwidth}
        \centering
        \includegraphics[width=\textwidth, keepaspectratio]{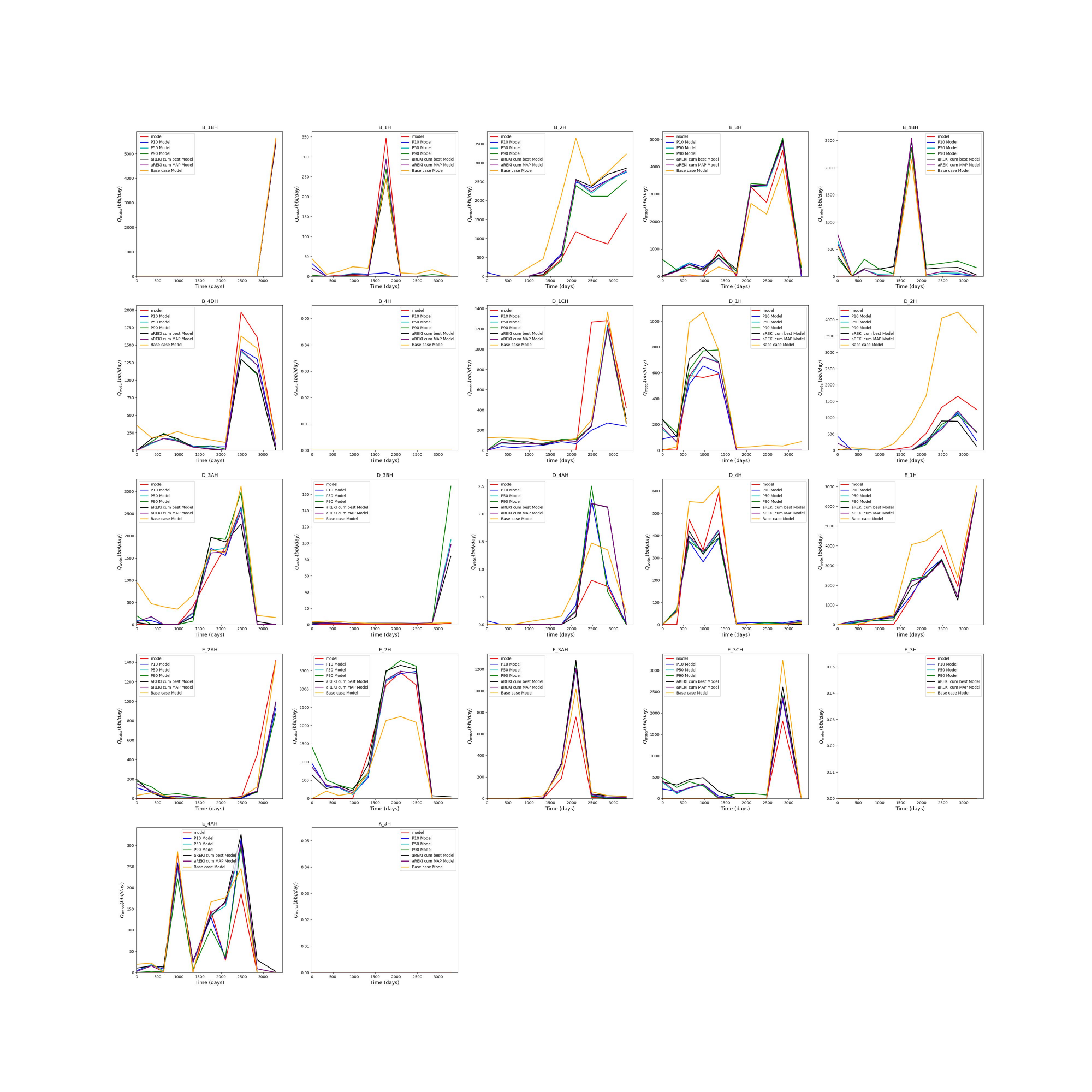}
        \caption{\textit{$Q_w\frac{(stb)}{(day)}$ }}
        \label{fig:water}
    \end{subfigure}    
    \vspace{0.5cm}    
    \begin{subfigure}[b]{0.45\textwidth}
        \centering
        \includegraphics[width=\textwidth, keepaspectratio]{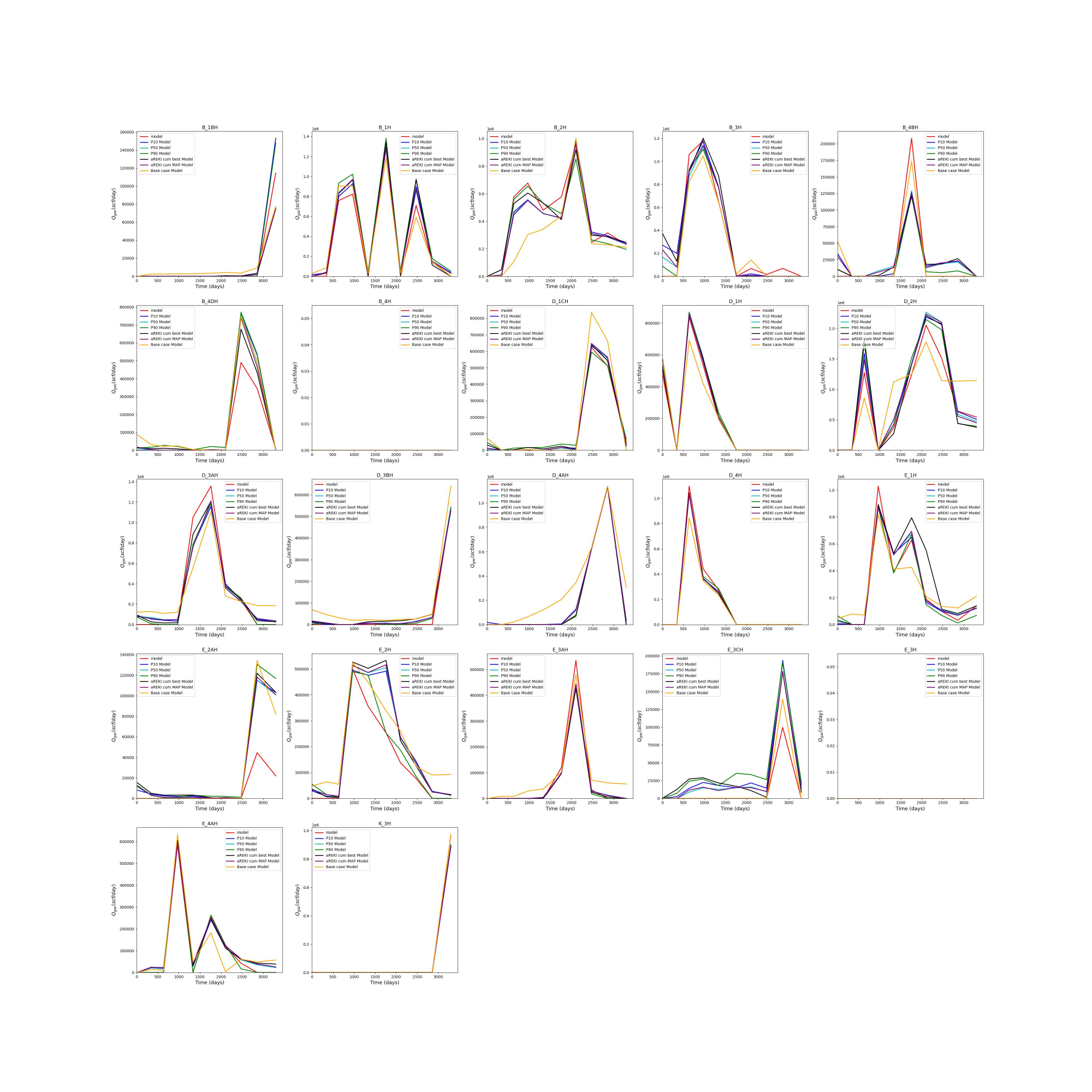}
        \caption{\textit{ $Q_g\frac{(scf)}{(day)}$ }}
        \label{fig:gas}
    \end{subfigure}
    \hfill
    \begin{subfigure}[b]{0.45\textwidth}
        \centering
        \includegraphics[width=\textwidth, keepaspectratio]{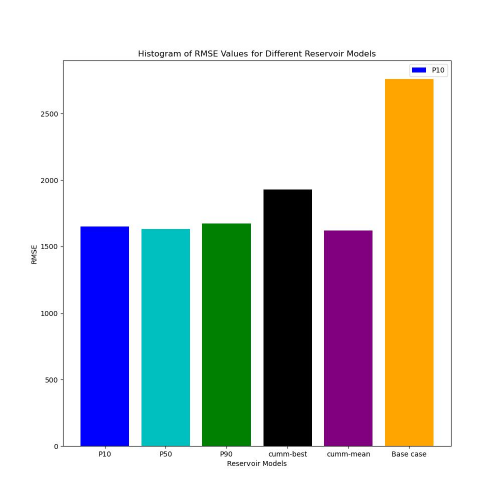} 
        \caption{\textit{ RMSE value showing the performance of the 6-reservoir model accuracy to the True data to be matched.}}
        \label{fig:newimage}
    \end{subfigure}    
    \caption{\textit{P10-P50-P90 Production profile comparison for the posterior ensemble. (red) True model, (Blue) P10 model, (cyan) P50 model, (green) P90 model.  from the 22 producers. They are 22 oil/water/gas producers, 9 water injectors and 4 gas injectors.}}
    \label{fig:all_images}
\end{figure}
\section*{Conclusion}
We have developed a coupled mixture of experts -  physics-informed neural operator-based reservoir simulator (\textit{PINO-CCR})  that solves the black oil model by augmenting the supervised loss from the labelled data with the residual loss of the governing pde. The \textit{PINO} can approximate the pressure, water saturation and gas saturation given any input of the permeability , porosity and the fault multiplier term . The developed \textit{PINO-CCR} is first compared to the numerical black oil reservoir simulator and the similarities are very close. Next, we use this developed surrogate simulator in an ensemble-based history-matching workflow. The method used is the \textit{a}REKI with VCAE and DDIM generative exotic priors, to make the inverse problem less ill-posed, reduce the computational complexity of the Kalman gain and improve convergence to the minimum of the history matching cost functional. For the \textit{Norne} field the overall workflow is successful in recovering the unknown permeability, porosity fields as well as the fault multiplier values and simulation (inference) is very fast with  an inference time of  approximately 7secs and a speedup of 6000X to the traditional numerical methods. Training of the \textit{PINO-CCR} surrogate for the \textit{Norne} field on a cluster with 2 NVIDIA H100 GPUs with 80G memory each takes approximately 5 hours for ~100 training samples. Full reservoir characterization together with plotting and solution to the inverse problem for an ensemble size of 200 realizations took ~1 hour(s). The workflow is suitable to be used in an inverse uncertainty quantification scenario where sampling the full posterior density is desirable

\section*{Credit authorship contribution statement}
Clement Etienam: Data curation, Conceptualization, Formal analysis, Investigation, Methodology, Software, Validation, Writing – original draft. Yang Juntao/Oleg Ovcharenko: Investigation, Writing – review and  editing – original draft. Issam Said: Research direction, conceptualisation of the forward and inverse problem. 

\section*{Acknowledgement}
The authors would like to express gratitude to NVIDIA for their financial support and permission to publish the article. Additionally, we extend our thanks to colleagues who contributed to this paper by way of intellectual discussions and suggestions.

\bibliographystyle{unsrt}  

\begin{thebibliography}{10}

\bibitem{emerick2013ensemble}
A. A. Emerick and A. C. Reynolds.
\newblock Ensemble smoother with multiple data assimilation.
\newblock {\em Computers Geosciences}, 2013.

\bibitem{Alakeely2022}
A. Alakeely, R. Horne
\newblock 2022. Simulating oil and water production in reservoirs with generative deep learning. 
\newblock SPE Reservoir Eval. Eng. 25 (4), 751e773. https://doi.org/10.2118/206126-PA.

\bibitem{Erofeev2019}
A. Erofeev, D. Orlov, A. Ryzhov, D. Koroteev
\newblock 2019. Prediction of porosity and permeability alteration based on machine learning algorithms. 
\newblock Transp Porous Med. 128, 677e700. https://doi.org/10.1007/s11242-019-01265-3.


\bibitem{stuart2010inverse}
A. M. Stuart.
\newblock Inverse problems: A Bayesian perspective.
\newblock {\em Acta Numerica}, 19:451--559, 2010.

\bibitem{tarantola2005inverse}
A. Tarantola.
\newblock Inverse problem theory and methods for model parameter estimation.
\newblock SIAM, Philadelphia, 2005.

\bibitem{Norne2}
B. Foss, , and J.L, Jensen. (2002). 
\newblock Stochastic reservoir simulation for the modelling of uncertainty in the Norne field. 
\newblock SPE Reservoir Evaluation and Engineering.

\bibitem{burt2020}
D.R., Burt, C.E.,Rasmussen, and  M., van der Wilk. 
\newblock Convergence of sparse variational inference in Gaussian processes regression, 2020 
\newblock Journal of Machine Learning Research, 21:1–63

\bibitem{bishop2006pattern}
C. M. Bishop.
\newblock Pattern Recognition and Machine Learning (Information Science and Statistics).
\newblock Springer-Verlag, Berlin, Heidelberg, 2006.

\bibitem{etienam2019seismic}
C. Etienam.
\newblock 4D Seismic History Matching Incorporating Unsupervised Learning.
\newblock {\em Society of Petroleum Engineers B}, 2019.

\bibitem{etienam2023ultrafast}
C. Etienam, J. H. L. Law, and S. Wade.
\newblock Ultra-fast Deep Mixtures of Gaussian Process Experts.
\newblock 2023.
\newblock \url{https://arxiv.org/abs/2006.13309}.

\bibitem{etienam2023reservoir}
C. Etienam, I. Said, O. Ovcharenko, and K. Hester.
\newblock A Reservoir Model Characterization with a Bayesian Framework and a Modulus-based Physics-constrained Neural Operator.
\newblock In {\em EAGE Seventh High-Performance Computing Workshop}, 2023.
\newblock \doi{10.3997/2214-4609.2023630004}.

\bibitem{etienam2024novel}
C. Etienam, Y. Juntao, I. Said, O. Ovcharenko, K. Tangsali, P. Dimitrov, K. Hester
\newblock 2024. A Novel A.I Enhanced Reservoir Characterization with a Combined Mixture of Experts -- NVIDIA Modulus based Physics Informed Neural Operator Forward Model
\newblock \url{https://arXiv:2404.14447}.

\bibitem{CarlRass2010}
C. E. Rasmussen and H. Nickisch.
\newblock Gaussian processes for machine learning {(GPML)} toolbox.
\newblock {\em Journal of Machine Learning Research}, 11:3011--3015, 2010.

\bibitem{Gasmi2021}
C.F. Gasmi, H. Tchelepi
\newblock 2021. Physics Informed Deep Learning for Flow and Transport in Porous Media 
\newblock \url{https:/arXiv:2104.02629}.

\bibitem{oliver2010recent}
D. S. Oliver and Y. Chen.
\newblock Recent progress on reservoir history matching: a review.
\newblock 2010.

\bibitem{oliver2008inverse}
D. S. Oliver, A. C. Reynolds, and N. Liu.
\newblock Inverse Theory for Petroleum Reservoir Characterization and History Matching.
\newblock Cambridge University Press, 2008.

\bibitem{bernholdt2019cluster}
D. E. Bernholdt, M. R. Cianciosa, D. L. Green, J. M. Park, K. J. H. Law, and C. Etienam.
\newblock Cluster, classify, regress: A general method for learning discontinuous functions.
\newblock {\em Foundations of Data Science}, 1(2639-8001-2019-4-491):491, 2019.

\bibitem{kingma2014autoencoding}
D, Kingma and M. Welling.
\newblock  Auto-Encoding Variational Bayes
\newblock {\em 2nd International Conference on Learning Representations, ICLR 2014}:2014.
\newblock \url{https://arxiv.org/abs/1312.6114}.

\bibitem{otter2021}
D.W. Otter, J.R. Medina, J.K. Kalita. 
\newblock 2021. A survey of the usages of deep learning for natural language processing. 
\newblock IEEE Transact. Neural Networks Learn. Syst. 32, 604e624. https://doi.org/10.1109/TNNLS.2020.2979670.

\bibitem{snelson2006}
E. Snelson and Z. Ghahramani.
\newblock Sparse {Gaussian} processes using pseudo-inputs.
\newblock In {\em Advances in Neural Information Processing Systems}, pages 1257--1264, 2006.

\bibitem{evensen2003ensemble}
G. Evensen.
\newblock The Ensemble Kalman Filter: Theoretical formulation and practical implementation.
\newblock {\em Ocean Dynamics}, 53:343--367, 2003.

\bibitem{Kania2021}
G.E. Karniadakis, I.G. Kevrekidis, L. Lu, P. Perdikaris, S. Wang, L. Yang
\newblock 2021. Physics-informed machine learning. 
\newblock Nat. Rev. Phys. 3, 422e440. https://doi.org/10.1038/s42254-021-00314-5.

\bibitem{goodfellow2014generative}
I. Goodfellow, J. Pouget-Abadie, M. Mirza, B. Xu, D. Warde-Farley, S. Ozair, A. Courville, and Y. Bengio.
\newblock Generative adversarial nets.
\newblock {\em Advances in Neural Information Processing Systems}, 2014:2672--2680.
\newblock \url{http://papers.nips.cc/paper/5423-generative-adversarial-nets.pdf}.


\bibitem{nocedal1999numerical}
J. Nocedal and S. J. Wright.
\newblock Numerical Optimization.
\newblock Springer, New York, 1999.

\bibitem{song2020denoising}
J. Song, C. Meng, and S.Ermon
\newblock 2020. Denoising Diffusion Implicit Models
\newblock \url{https://arXiv:2010.02502}.

\bibitem{sherman1950adjustment}
J. Sherman and W. J. Morrison.
\newblock Adjustment of an inverse matrix corresponding to a change in one element of a given matrix.
\newblock {\em The Annals of Mathematical Statistics}, 21:124--127, 1950.

\bibitem{adler2017solving}
J. Adler and O. Oktem.
\newblock Solving ill-posed inverse problems using iterative deep neural networks.
\newblock {\em Inverse Problems}, 33:124007, 2017.
\newblock \url{http://stacks.iop.org/0266-5611/33/i=12/a=124007}.

\bibitem{zhu2017toward}
J.-Y. Zhu, R. Zhang, D. Pathak, T. Darrell, A. A. Efros, O. Wang, and E. Shechtman.
\newblock Toward multimodal image-to-image translation.
\newblock {\em Advances in Neural Information Processing Systems}, 2017:465--476.
\newblock \url{https://papers.nips.cc/paper/7189-toward-multimodal-image-to-image-translation.pdf}.

\bibitem{hanna2022}
J.M. Hanna, J.V. Aguado, S. Comas-Cardona, R. Askri, D. Borzacchiello
\newblock 2022. Residual-based adaptivity for two-phase flow simulation in porous media using physics-informed neural networks. 
\newblock Comput. Methods Appl. Mech. Eng. 396, 115100. https://doi.org/10.1016/j.cma.2022.115100.

\bibitem{cornelio2022}
J. Cornelio, S.M. Razak, Y. Cho, H. Liu, R. Vaidya, B. Jafarpour
\newblock 2022. Residual learning to integrate neural network and physics-based models for improved production prediction in unconventional reservoirs. 
\newblock SPE J. 27 (6), 3328e3350.\url{https://doi.org/10.2118/210559-PA}.

\bibitem{Li2022}
J. Li, D. Zhang, N. Wang, H. Chang
\newblock 2022. Deep learning of two-phase flow in porous media via theory-guided neural networks. 
\newblock SPE J. 27 (2), 1176e1194. \url{https://doi.org/10.2118/208602-PA}.

\bibitem{santos2021}
J.E. Santos, Y. Yin, H. Jo, W. Pan, Q. Kang, H.S. Viswanathan, M. Prodanovic, M.J. Pyrcz, N. Lubbers
\newblock 2021. Computationally efficient multiscale neural networks applied to fluid flow in complex 3D porous media. 
\newblock Transp Porous Med. 140, 241e272. https://doi.org/10.1007/s11242-021-01617-y.

\bibitem{kim2021}
J. Kim, C. Park, S. Ahn, B. Kang, H. Jung, I. Jang
\newblock 2021. Iterative learning-based many-objective history matching using deep neural network with stacked autoencoder. 
\newblock Petrol. j.petsci.2021.08.001.

\bibitem{gardner2021gpytorch}
J. R. Gardner , G. Pleiss, D. Bindel, K. Q. Weinberger, A. G, Wilso
\newblock GPyTorch: Blackbox Matrix-Matrix Gaussian Process Inference with GPU Acceleration,2021
\newblock  \url{https://arxiv.org/abs/1809.11165}

\bibitem{quinonero2005}
J. Qui{\~n}onero-Candela and C. E. Rasmussen.
\newblock A unifying view of sparse approximate {Gaussian} process regression.
\newblock {\em Journal of Machine Learning Research}, 6:1939--1959, 2005.

\bibitem{law2012evaluating}
K. J. H. Law and A. M. Stuart.
\newblock Evaluating Data Assimilation Algorithms.
\newblock {\em Mon. Weather Rev.}, 140:37--57, 2012.

\bibitem{yang2019physics}
L. Yang, D. Zhang, and G. E. Karniadakis.
\newblock Physics-informed generative adversarial networks for stochastic differential equations.
\newblock 2019.
\newblock arXiv preprint arXiv:1811.02033.

\bibitem{Daolun2021}
L. Daolun, S. Luhang, Z. Wenshu, L. Xuliang, T. Jieqing
\newblock 2021. Physics-constrained deep learning for solving seepage equation. 
\newblock J. Petrol. Sci. Eng. 206, 109046. https://doi.org/10.1016/j.petrol.2021.109046.

\bibitem{hanke1997regularizing}
M. Hanke.
\newblock A regularizing Levenberg-Marquardt scheme, with applications to inverse groundwater filtration problems.
\newblock {\em Inverse Problems}, 13:79--95, 1997.

\bibitem{Alma2022}
M.M. Almajid, M.O. Abu-Al-Saud
\newblock 2022. Prediction of porous media fluid flow using physics informed neural networks. 
\newblock J. Petrol. Sci. Eng. 208, 109205. https://doi.org/10.1016/j.petrol.2021.109205.

\bibitem{iglesias2013regularizing}
M. A. Iglesias and C. Dawson.
\newblock The regularizing Levenberg-Marquardt scheme for history matching of petroleum reservoirs.
\newblock {\em Computational Geosciences}, 17:1033--1053, 2013.

\bibitem{raissi2019physics}
M. Raissi, P. Perdikaris, and G. E. Karniadakis.
\newblock Physics-informed neural networks: A deep learning framework for solving forward and inverse problems involving nonlinear partial differential equations.
\newblock {\em Journal of Computational Physics}, 378:686--707, 2019.
\newblock \doi{10.1016/j.jcp.2018.10.045}.
\newblock \url{http://www.sciencedirect.com/science/article/pii/S0021999118307125}.

\bibitem{raissi2019forward}
M. Raissi.
\newblock Forward-backward stochastic neural networks: Deep learning of high-dimensional partial differential equations.
\newblock 2019.
\newblock arXiv preprint arXiv:1804.07010.

\bibitem{raissi2017physics}
M. Raissi, P. Perdikaris, and G. E. Karniadakis.
\newblock Physics Informed Deep Learning (Part I): Data-driven solutions of nonlinear partial differential equations.
\newblock 2017.
\newblock arXiv preprint arXiv:1711.10561.

\bibitem{trapp2020deep}
M. Trapp, R. Peharz, F. Pernkopf, and C. E. Rasmussen.
\newblock Deep Structured Mixtures of Gaussian Processes.
\newblock In {\em International Conference on Artificial Intelligence and Statistics}, pages 2251--2261. PMLR, 2020.

\bibitem{bauer2016}
M. Bauer, M. van der Wilk, and C. E. Rasmussen.
\newblock Understanding Probabilistic Sparse Gaussian Process Approximations.
\newblock In {\em Advances in Neural Information Processing Systems 29}, edited by D. D. Lee, M. Sugiyama, U. V. Luxburg, I. Guyon, and R. Garnett, pages 1533--1541. Curran Associates, Inc., 2016.
\newblock \url{http://papers.nips.cc/paper/6477-understanding-probabilistic-sparse-gaussian-process-approximations.pdf}.

\bibitem{titsias2009}
M. Titsias.
\newblock Variational learning of inducing variables in {sparse Gaussian} processes.
\newblock In {\em Artificial Intelligence and Statistics}, pages 567--574, 2009.

\bibitem{wang2022}
N. Wang, H. Chang, D. Zhang
\newblock 2022. Surrogate and inverse modeling for two-phase flow in porous media via theory-guided convolutional neural network. 
\newblock J. Comput. Phys. 466, 111419. https://doi.org/10.1016/j.jcp.2022.111419.

\bibitem{Modulus}
NVIDIA Modulus Team. NVIDIA Modulus [Computer software]. 
\newblock https://github.com/NVIDIA/modulus/tree/main

\bibitem{dorn2008history2}
O. Dorn and R. Villegas.
\newblock History matching of petroleum reservoirs using a level set technique.
\newblock {\em Inverse Problems}, 24(3):035015, 2008.

\bibitem{Dong2019}
P. Dong, X. Liao, Z. Chen, H. Chu. 
\newblock 2019. An improved method for predicting CO2 minimum miscibility pressure based on artificial neural network. 
\newblock Adv. GeoEnergy Res. 3, 355e364. https://doi.org/10.26804/ager.2019.04.02.

\bibitem{OPM}
R. Blöchliger, M. Blatt, A. Bürger, H. Hajibeygi, R. Klöfkorn, and O. Møyner. 
\newblock The OPM Flow Simulator for Geologic Carbon Storage. 
\newblock Computers and Geosciences, 2020

\bibitem{aanonsen2009ensemble}
S. Aanonsen, D. Oliver, A. Reynolds, and B. Valles.
\newblock The Ensemble Kalman Filter in Reservoir Engineering--a Review.
\newblock {\em SPE Journal}, 14(11):393--412, 2009.


\bibitem{rojas1998nonlinear}
S. Rojas and J. Koplik.
\newblock Nonlinear numerical solver in porous media.
\newblock {\em Phys. Rev. E}, 58:4776--4782, 1998.
\newblock \doi{10.1103/PhysRevE.58.4776}.
\newblock \url{https://link.aps.org/doi/10.1103/PhysRevE.58.4776}.

\bibitem{yuksel2012twenty}
S. E. Yuksel, J. N. Wilson, and P. D. Gader.
\newblock Twenty years of mixture of experts.
\newblock {\em IEEE transactions on neural networks and learning systems}, 23(8):1177--1193, 2012.

\bibitem{Moosavi2019}
S.R. Moosavi, B. Vaferi, D.A. Wood
\newblock 2020. Auto-detection interpretation model for horizontal oil wells using pressure transient responses. 
\newblock Adv. Geo-Energy Res. 4, 305e316. https://doi.org/10.46690/ager.2020.03.08.


\bibitem{bui2016}
T. D. Bui, J. Yan, and R. E. Turner.
\newblock A unifying framework for sparse {Gaussian} process approximation using power expectation propagation.
\newblock {\em Journal of Machine Learning Research}, 18:1--72, 2017.

\bibitem{Ertekin2019}
T. Ertekin, Q. Sun. 
\newblock 2019. Artificial intelligence applications in reservoir engineering: a status check. 
\newblock Energies 12, 2897. https://doi.org/10.3390/en12152897.

\bibitem{chen2016xgboost}
T. Chen, T., and  C. Guestrin, C. 
\newblock \textit{XGBoost: A scalable tree boosting system}. (2016)
\newblock Proceedings of the 22nd ACM SIGKDD International Conference on Knowledge Discovery and Data Mining, 785--794.

\bibitem{chung2020}
T. Chung, Y. Da Wang, R.T. Armstrong, P. Mostaghimi
\newblock 2020. CNN-PFVS: integrating neural network and finite volume models to accelerate flow simulation on pore space images. 
\newblock Transp Porous Med.135, 25e37. https://doi.org/10.1007/s11242-020-01466-1.


\bibitem{trespBCM}
V. Tresp.
\newblock A Bayesian committee machine.
\newblock {\em Neural computation}, 12(11):2719--2741, 2000.


\bibitem{Lecun2015}
Y. LeCun, Y. Bengio, G. Hinton. 
\newblock 2015. Deep learning. Nature 521, 436e444. 
\newblock https:// doi.org/10.1038/nature14539.

\bibitem{chen2011ensemble}
Y. Chen and D. Oliver.
\newblock Ensemble randomized maximum likelihood method as an iterative ensemble smoother.
\newblock {\em Mathematical Geosciences}, 2011.

\bibitem{Norne1}
Y. Chen, and D. S. Oliver. (2014). 
\newblock Ensemble-based methods for solving reservoir engineering problems applied to Norne field data. 
\newblock Energy Exploration and Exploitation.

\bibitem{li2023physics}
Z. Li, H. Zheng, N. Kovachki, D. Jin, H. Chen, B. Liu, K. Azizzadenesheli, and A. Anandkumar.
\newblock Physics-Informed Neural Operator for Learning Partial Differential Equations.
\newblock {\em Journal Name or Conference Proceedings}, 2023.
\newblock \url{https://arxiv.org/pdf/2111.03794.pdf}.

\bibitem{li2021fourier}
Z. Li, N. Kovachki, K. Azizzadenesheli, B. Liu, K. Bhattacharya, A. Stuart, and A. Anandkumar.
\newblock Fourier Neural Operator for Parametric Partial Differential Equations.
\newblock {\em Journal Name or Conference Proceedings}, 2021.
\newblock \url{https://arxiv.org/abs/2010.08895}.

\end{thebibliography}

\end{document}